\documentclass[sigconf]{acmart}

\usepackage{booktabs}    
\usepackage{mathtools}
\usepackage{multirow}
\usepackage{subcaption}
\usepackage{tabularx}
\usepackage{amsmath,bm}
\usepackage{algorithm}
\usepackage{algpseudocode}
\usepackage{wrapfig}
\usepackage{lipsum}
\usepackage{float}
\usepackage{arydshln}

\algnewcommand{\Inputs}[1]{%
  \State \textbf{Inputs:} {\raggedright #1}
}
\algnewcommand{\Initialize}[1]{%
  \State \textbf{Initialize:} {\raggedright #1}
}

\setcopyright{rightsretained}

\acmYear{2023}

\begin{document}

\title{Discovering Attention-Based Genetic Algorithms via Meta-Black-Box Optimization}

\author{Robert Tjarko Lange*}\thanks{* Work done as interns at DeepMind. Contact: \texttt{robert.t.lange@tu-berlin.de}.}
\affiliation{%
    \institution{Technical University Berlin}
}

\author{Tom Schaul}
\affiliation{%
    \institution{DeepMind}
}

\author{Yutian Chen}
\affiliation{%
    \institution{DeepMind}
}

\author{Chris Lu*}
\affiliation{%
    \institution{University of Oxford}
}

\author{Tom Zahavy}
\affiliation{%
    \institution{DeepMind}
}

\author{Valentin Dalibard}
\affiliation{%
    \institution{DeepMind}
}

\author{Sebastian Flennerhag}
\affiliation{%
    \institution{DeepMind}
}

\renewcommand{\shortauthors}{Lange et al.}

\begin{abstract}
Genetic algorithms constitute a family of black-box optimization algorithms, which take inspiration from the principles of biological evolution. While they provide a general-purpose tool for optimization, their particular instantiations can be heuristic and motivated by loose biological intuition. In this work we explore a fundamentally different approach: Given a sufficiently flexible parametrization of the genetic operators, we discover entirely new genetic algorithms in a data-driven fashion. More specifically, we parametrize selection and mutation rate adaptation as cross- and self-attention modules and use \emph{Meta-Black-Box-Optimization} to evolve their parameters on a set of diverse optimization tasks. The resulting \emph{Learned Genetic Algorithm} outperforms state-of-the-art adaptive baseline genetic algorithms and generalizes far beyond its meta-training settings. The learned algorithm can be applied to previously unseen optimization problems, search dimensions \& evaluation budgets. We conduct extensive analysis of the discovered operators and provide ablation experiments, which highlight the benefits of flexible module parametrization and the ability to transfer (`plug-in') the learned operators to conventional genetic algorithms.
\end{abstract}

%
%



\begin{teaserfigure}
  \centering
  \includegraphics[width=0.95\textwidth]{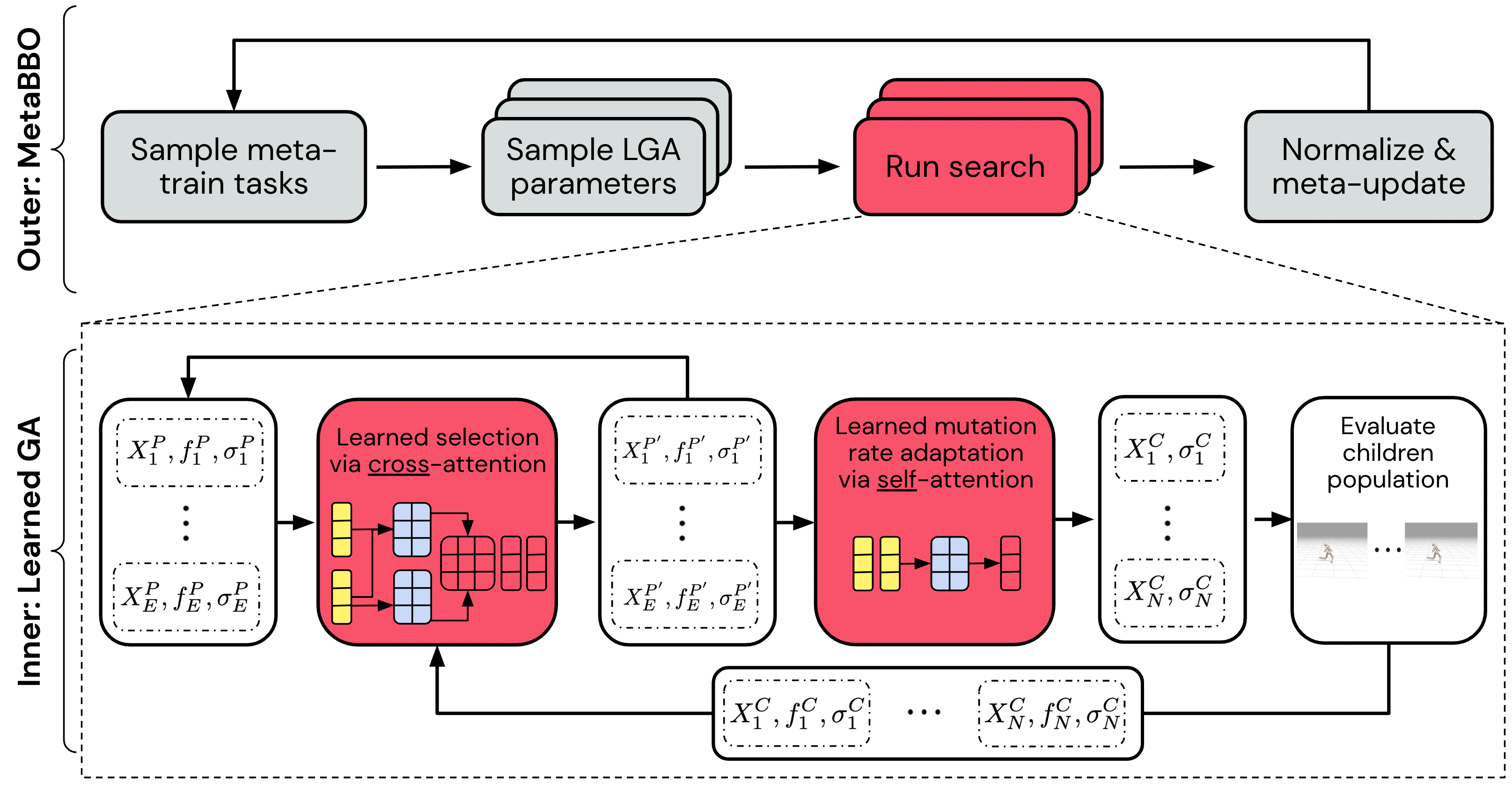}
  \caption{Discovering attention-based Learned Genetic Algorithms (LGA) via MetaBBO. At each meta-iteration one samples a set of inner loop tasks and a set of candidate LGA parameters from a meta-evolutionary optimizers (EO). Afterwards, one runs an inner loop search and compute a normalized meta-fitness score across the tasks. We update the meta-EO and iterate.}
  \label{fig:lga}
\end{teaserfigure}

\maketitle

\newpage
\section{Introduction}
\label{sec:introduction}

\textbf{Motivation}. Genetic algorithms (GAs) provide a set of evolution-inspired optimization algorithms, which are flexibly applicable to black-box optimization (BBO) problems. They commonly rely on human designed operators, which impose a restrictive and most importantly \textit{subjective} set of manual priors. This bears the risk of domain overfitting and limited generalization capabilities. Based on recent results in the discovery of attention-based Evolution Strategies \citep{les2022}, we propose that these limitations can be overcome by meta-learning effective GA operators from data. Thereby, the inductive biases of the GA itself are indirectly encoded by its parametrization \& can be discovered in an optimization-driven fashion, i.e. by improving its meta-performance on a distribution of relevant tasks. \\
\textbf{Approach}. Inspired by the recent success of the Set Transformer \citep{lee2019set} architecture, we introduce neural network-based architectures to substitute core genetic operators: \textit{Selection} and \textit{mutation rate adaptation} are cast as dot-product attention modules, which can flexibly be applied to problems with varying dimensions \& population sizes. The resulting family of genetic algorithms can implement different operations based on the specific module weights. We meta-evolve these weights on a set of representative optimization problems using meta-black-box optimization (\emph{MetaBBO}, \citep{les2022}). \\
\textbf{Results}. We evaluate the performance and generalization capabilities of the meta-trained Learned Genetic Algorithm (\emph{LGA}). LGA is capable of generalizing far beyond its meta-training distribution and outperforms established baseline GAs on several BBO benchmarks (BBOB) \citep{hansen2010real, arango2021hpo} and neuroevolution problems. This includes different optimization functions, number of search dimensions and population sizes. Furthermore, we analyze the discovered neural network GA operators: The selection operator has learned an adaptive form of truncation, which maintains diversity and redundancy among the parent population.
The learned mutation rate adaption operator, on the other hand, automatically scales the amount of exploration in a task-dependent fashion.
We investigate the importance of the meta-evolution task distribution: While it is possible to meta-evolve effective LGAs on as little as five BBOB functions, we show that LGAs can overfit their meta-training distribution and one has to ensure sufficient meta-regularization for broad generalization.
Trained LGAs are robust to their choice of hyperparameters and the details of their meta-training procedure. Finally, we show that the individual learned operators can replace white-box GA operators inducing a positive transfer effect. 
\\
\textbf{Contributions}. Our contributions are summarized as follows:
\begin{enumerate}
    \item We propose a dot-product attention-based parametrization of the selection \& mutation rate adaptation GA operators.
    \item We discover new GAs by meta-evolving their parameters based on the performance on meta-training BBO tasks.
    \item The resulting LGA is capable of generalizing far beyond its meta-training settings and outperforms several GA baselines on various benchmark tasks and evaluation budgets.
    \item We perform several ablations to LGA to assess the contributions of learning the individual operator modules. Both learned operators contribute to the overall performance. 
    \item We highlight the robustness of the trained LGA to its hyperparameters, the transferability of the learned components and the stability of the MetaBBO procedure.
\end{enumerate}
%

\newpage
\section{Related Work}
\label{sec:related}
\textbf{Discovery via Meta-Learned Algorithms}. Recent efforts have proposed to replace manually designed algorithms by end-to-end optimized inductive biases, by meta-learning parametrized components on a representative task distribution. E.g. this includes the discovery of Reinforcement Learning objectives \citep{oh2020discovering, xu2020meta, lu2022discovered}, schedules of algorithm hyperparameters via meta-gradients \citep{xu2018meta, zahavy2020self, flennerhag2021bootstrapped, parker2022automated}, and the meta-learning of entire learning algorithms \citep{wang2016learning, kirsch2021meta, kirsch2022introducing}. The discovery process can be supported by suitable neural network architectures. Our proposed LGA architecture leverages attention layers to derive a neural network-based family of GAs.\\
\textbf{Meta-Learned Gradient-Based Optimization}. Our work is closely related to the ambition of meta-learning gradient descent-based learning rules \citep{bengio_1992, andrychowicz_2016, metz2019understanding}. These approaches rely on access to efficient gradient calculations via the backpropagation algorithm and thereby do not apply to BBO problems. A small neural network processes the gradient and standard optimizer statistics (momentum, etc.) to output a weight change. The optimizer network weights in turn have been meta-learned on a task distribution \citep{metz_2020tasks}. \citet{metz2022velo} showed that this results in a highly competitive optimizer for deep learning tasks. Our MetaBBO-discovered LGA, on the other hand, provides a general-purpose BBO, which does not require differentiable objective functions.\\
\textbf{Meta-Learned Population-Based Optimization}. \citet{shala2020learning} meta-learn a controller for the scalar mutation rate in CMA-ES \citep{hansen2001completely}. \citet{chen_2017,tv2019meta, gomes2021meta} previously optimized entire neural network-parametrized algorithms for low-dimensional BBO. All of them use a recurrent network, which processes raw solution candidates and their respective fitness scores. These methods often struggle to generalize to new optimization domains and are often constrained to fixed population sizes and/or search dimensions.
\citet{les2022} recently leveraged the equivariance property of dot-product self-attention to the input ordering \citep{lee2019set, kossen2021self, tang2021sensory} to learn adaptive recombination weights for evolution strategies.
The proposed LGA extends this attention-based BBO perspective in order to characterize GA operators. After successful meta-training, the learned GA is capable of generalizing to unseen populations and large search spaces. To the best of our knowledge we are the first to demonstrate that a meta-learned GA generalizes to challenging neuroevolution tasks. Finally, the MetaBBO approach does not require access to knowledge of meta-task optima \citep{tv2019meta} or a teacher algorithm \citep{shala2020learning}.\\
\textbf{Baseline Genetic Algorithms}. Throughout the paper we compare against four competitive baseline GAs including the following:
\begin{itemize}
    \item Gaussian GA \citep{rechenberg1973evolutionsstrategie}: A simple GA with Gaussian perturbations and fixed mutation strength using truncation selection.
    \item MR-1/5 GA \citep{rechenberg1973evolutionsstrategie}: Doubles the mutation rate if 1/5 of all perturbations were beneficial. Otherwise, the rate is halved.
    \item SAMR-GA \citep{clune2008natural}: Self-adapts per-parent mutation rates based on a simple co-evolution heuristic and meta-mutation rate.
    \item GESMR-GA \citep{kumar2022effective}: Avoids vanishing mutation rates by using a group elite selection criterion and mutation rate sharing.
\end{itemize}
We additionally consider Sep-CMA-ES \citep{ros2008simple} as a scalable evolution strategy baseline for neuroevolution tasks. Each baseline GA was tuned using small grid search sweeps (see Appendix \ref{appendix:settings}). Otherwise, we adopted the settings provided by the authors.

\newpage
\section{Background}
\label{sec:background}

\textbf{Black-Box Optimization}. Throughout this manuscript, we are interested in efficient continuous black-box optimization: Given a function $f(x): \mathbb{R}^D \to \mathbb{R}$ with unknown functional form, i.e. we cannot compute its derivative, we seek to find its global optimum:

\vspace{-0.25cm}
$$\min_\mathbf{x} f(\mathbf{x}), \text{s.t.} \ \mathbf{x}_d \in [l_d, u_d] \subset [-\infty, \infty], \forall d=1,...,D,$$
\textbf{Genetic Algorithms}. GAs provide a class of BBO algorithms, which iteratively evaluate a population consisting of $N$ solution candidates $X^C = [\mathbf{x}^C_1, \dots, \mathbf{x}^C_N]^T \in \mathbb{R}^{N \times D}$ (`children') with fitness $\mathbf{f}^C \in \mathbb{R}^N$. Given a set of $E$ `parent' solutions $X^P = [\mathbf{x}^P_1, \dots, \mathbf{x}^P_{E}]^T \in \mathbb{R}^{E \times D}$ with associated fitness $\mathbf{f}^{P} \in \mathbb{R}^E$, the parents are replaced by the children using a heuristic fitness-based selection criterion.\\
Most GAs make use of truncation selection in which all children and parents $[\mathbf{x}^C_1, \dots, \mathbf{x}^C_N, \mathbf{x}^P_1, \dots, \mathbf{x}^P_E]^T$ are jointly sorted by their fitness. The top-$E$ performing solutions replace the parent archive:

\vspace{-0.25cm}
$$X^{P'}, \mathbf{f}^{P'} = \texttt{Selection}(X^P, \mathbf{f}^P, X^C, \mathbf{f}^C).$$
Commonly the number of parents is set to be small $E \ll N$ and often even $E=1$, enforcing a type of hill climbing.
%
%
%
%
$N$ children candidates are uniformly sampled with replacement from the parents: 

\vspace{-0.25cm}
$$\tilde{X}^P, \tilde{\mathbf{f}}^P = \texttt{Sample}(X^P, \mathbf{f}^P) \in \mathbb{R}^{N \times D}.$$
Afterwards, they are perturbed using a mutation rate (MR) $\sigma \in \mathbb{R}_+$, which controls the strength of the Gaussian noise, $\epsilon_j \sim \mathcal{N}(\mathbf{0}_D, \mathbf{I}_D)$:
%
$$X_j^C = \texttt{Mutation}(\tilde{X}^P_j, \sigma) = \tilde{X}^P_j + \sigma \epsilon_j \in \mathbb{R}^{D}, \ \forall j=1,...,N.$$
Many competitive GAs keep a vector of parent-specific \citep{clune2008natural} mutation rates $\boldsymbol{\sigma}^P \in \mathbb{R}^E$ and additionally perform an intermediate mutation rate adaptation (MRA) step to improve the mutation rate(s) given information gathered throughout the fitness evaluations:
%
$$\boldsymbol{\sigma}^{P} = \texttt{Adaptation}(\mathbf{f}^{P}, \boldsymbol{\sigma}^P, \mathbf{f}^{C}, \boldsymbol{\sigma}^C) \in \mathbb{R}^E.$$
In this case, the children are perturbed by their sampled individual-specific mutation rate ($\boldsymbol{\sigma}^C \leftarrow \tilde{\boldsymbol{\sigma}}^P \in \mathbb{R}^N$) and selection also applies to the children's MR. Alternatively, GESMR-GA \citep{kumar2022effective} forms parent sub-groups and online co-evolves group-level mutation rates based on their observed fitness improvements.\\
\textbf{Set Operations via Dot-Product Self-Attention}. Scaled dot-product attention (SDPA) is especially well suited to characterize algorithms performing set operations, since it naturally enforces a permutation invariant function. Consider the standard formulation of SDPA, which embeds a set of $N$ input tokens, $X \in \mathbb{R}^{N \times D}$, into $D_K$-dimensional latent query $Q$, key $K$ and value $V$ representations:
\vspace{-0.25cm}
\begin{align*}
    Q &= X W_{Q} \in \mathbb{R}^{N \times D_K}, \\
    K &= X W_{K} \in \mathbb{R}^{N \times D_K}, \\
    V &= X W_{V} \in \mathbb{R}^{N \times D_K}.
\end{align*}
The output $Y$ is computed as a linear combination of the values:
%
\begin{align*}
Y = \text{softmax}\left(Q K^T/\sqrt{D_K}\right) V \in \mathbb{R}^{N \times D_K}.
\end{align*}
It can be shown that this transformation is equivariant to the ordering of the tokens in $X$, i.e. permuting the rows of $X$ will apply the same permutation to the rows of $Y$ \citep{tang2021sensory, kossen2021self}. We will leverage this suitable inductive bias to characterize GAs, which inherently operate on sets of solution candidates and their fitness scores.

\section{Attention-Based Genetic Operators}
\label{sec:lga}

\begin{figure}
\vspace{-0.5cm}
\caption{Learned Genetic Operators. \underline{Top}: Cross-attention selection between parent \& children fitness features. \underline{Bottom}: MRA via self-attention on parent mutation \& fitness features.}
\centering
\includegraphics[width=0.475\textwidth]{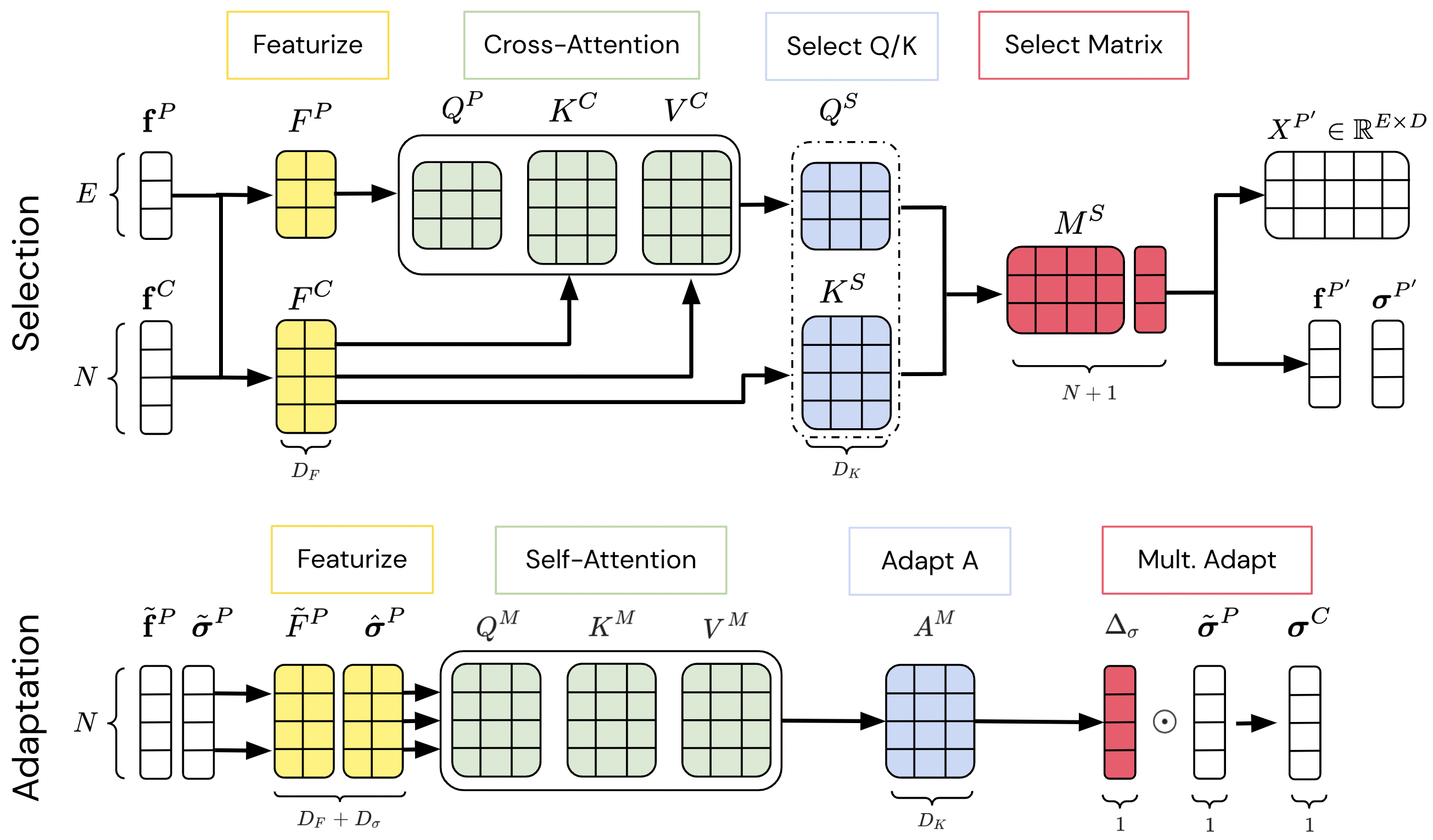}
\label{fig:learned_operators}
\vspace{-0.5cm}
\end{figure}

We now introduce an attention-based parametrization of the genetic $\texttt{Selection}$ and $\texttt{Adaptation}$ operators (Figure \ref{fig:learned_operators}). These in turn will be meta-optimized on a set of representative optimization tasks in order to capture useful BBO mechanisms.
We start by answering a natural question: What inputs should be processed by the operators in order to enable generalization across fitness and solution scales? \\
\textbf{Attention Features via Fitness Scores \& Mutation Strength}. To compute attention scores across parents and children, we need to construct sufficient features, which modulate effective selection and mutation rate adaptation. Furthermore, we want the meta-learned operations to generalize across different test optimization scales. Hence, we consider scale invariant normalizations: E.g. z-scoring and centered ranks (in $[-0.5, 0.5]$). We transform both the raw fitness scores ($D_F$ dim.) and the parent mutation rates ($D_\sigma$ dim.) to construct a set of features processed by the attention layers:

\begin{itemize}
    \item $F \in \mathbb{R}^{(N+E) \times D_F}$: Joint fitness transformations of parents and children (z-scores \& centered ranks).
    \item $F^C = F_{1:N}\in \mathbb{R}^{N \times D_F}$: Fitness transformations of children extracted from joint transforms (z-scores \& centered ranks).
    \item $F^P = F_{N:(N+E)}\in \mathbb{R}^{E \times D_F}$: Fitness transformations of parents extracted from joint transforms (z-scores \& centered ranks).
    \item $F^{P'} \in \mathbb{R}^{N \times D_F}$: (Separate) fitness transforms of sampled parents after selection operation (z-scores \& centered ranks).
    \item $\hat{\boldsymbol{\sigma}}^P \in \mathbb{R}^{N \times D_\sigma}$: Mutation strength transformations of sampled parents (z-scores \& [-1, 1] normalization).
    \item $F^M = [\tilde{F}^{P},\hat{\boldsymbol{\sigma}}^P] \in \mathbb{R}^{N \times (D_F + D_\sigma)}$: Concatenated fitness and mutation rate features of sampled parents.
\end{itemize}
The fitness features additionally include a Boolean indicating whether an individual performs better than the best fitness observed so far.\\
\textbf{Selection via Cross-Attention}. We replace the common sorting-based selection mechanism with a cross-SDPA layer, which compares children and parents. It first embeds the fitness transformations of the parents and children into queries, keys and values:
\vspace{-0.25cm}
\begin{align*}
    Q^P = F^P W_{Q^P} \in \mathbb{R}^{E \times D_K}, \\
    K^C = F^C W_{K^C}, \ V^C = F^C W_{V^C} \in \mathbb{R}^{N \times D_K}.
\end{align*}
Afterwards, we compute the normalized dot-product cross-attention features $A^S$ and construct a selection matrix $M^S \in \mathbb{R}^{E \times (N + 1)}$:

\begin{align*}
    A^S = \text{softmax}\left(\frac{Q^{P} (K^C)^T}{\sqrt{D_K}}\right) V^C \in \mathbb{R}^{E \times D_K} \\
    Q^S = A^S W_{Q^S} \in \mathbb{R}^{E \times D_k},\ K^S = F_C W_{K^S} \in \mathbb{R}^{N \times D_K} \\
    M^S = \text{softmax}\left(\left[\frac{Q^S (K^S)^T}{\sqrt{D_K}}, \mathbf{1}_E\right]\right) \in \mathbb{R}^{E \times (N + 1)}
\end{align*}
In the final line we concatenate an $E$-dimensional vector of ones to the outer product of $Q^S$ and $K^S$. Intuitively, this column represents a fixed offset used to indicate whether the parent copies any child at all or if it is not replaced. The rows of $M^S$ then specify the probability of each offspring to replace a parent:
\begin{align*}
M^S = \text{softmax}\left(\begin{bmatrix}
    m_{11} & m_{12} &  \dots  & m_{1N} & 1 \\
    m_{21} & m_{22} &  \dots  & m_{2N} & 1\\
    \vdots & \vdots &  \ddots & \vdots & 1\\
    m_{E1} & m_{E2} & \dots  & m_{EN} & 1
\end{bmatrix}\right)
\end{align*}
$M^S$ is row stochastic, i.e. the rows sum to 1. E.g. $M^S_{11}$ denotes the probability of replacing parent 1 with child 1, while $M^S_{1N+1}$ corresponds to not replacing the first parent.
We sample row-wise from a categorical distribution in order to determine whether a child replaces a particular parent. Afterwards, we use the selection matrix to update the parent archive ($X^{P'}$) and fitness archive ($\mathbf{f}^{P'}$):
$$ S \sim \text{Categorical}(M^S) \ \text{with } S \in \mathbb{R}^{E \times (N + 1)}, $$
$$ X^{P'} = X^C \cdot S_{:, 1:N} + X^P \cdot \text{diag}(S_{:, N:N+1}), $$
and similarly we obtain $\mathbf{f}^{P'}$ and $\boldsymbol{\sigma}^{P'}$ via masked addition. This selection operator can flexibly regulate the amount of truncation selection by replacing multiple parent slots with the same child.\\
\textbf{Mutation Rate Adaptation (MRA) via Self-Attention}. Next to selection we meta-learn MRA. The concatenated fitness and MR features of the sampled parents are processed by a SDPA layer, which outputs a child-specific feature matrix $A^M \in \mathbb{R}^{N \times D_K}$:
\begin{align*}
    K^M = F^M W_{K^M}, \ 
    Q^M = F^M W_{Q^M}, \
    V^M = F^M W_{V^M} \in \mathbb{R}^{N \times D_K},
\end{align*}
\begin{align*}
    A^M &= \text{softmax}\left(\frac{Q^M (K^M)^T}{\sqrt{D_K}}\right) V^M \in \mathbb{R}^{N \times D_K}.
\end{align*}
Afterwards, the multiplicative adaptation to the MR is constructed by projecting and re-parametrizing the attention output:
\begin{align*}
    \Delta_\sigma &= \exp(0.5 \times A^M W_\sigma) \in \mathbb{R}^{N}
\end{align*}
The children MR $\boldsymbol{\sigma}^C$ is obtained via element-wise multiplication:
\begin{align*}
    \boldsymbol{\sigma}^{C} &= \Delta_\sigma \odot \boldsymbol{\sigma}^{P'} \in \mathbb{R}^{N}
\end{align*}
$\theta = \{W_{Q^P}, W_{K^C}, W_{V^C}, W_{Q^S}, W_{K^S}, W_{Q^M}, W_{K^M}, W_{V^M}, W_\sigma\}$ denotes the joint attention weights, which characterize a specific instance of an LGA. We use a small feature dimension $D_K=16$, which results in <1500 trainable meta-parameters.
In summary, we introduced two dot-product attention-based operators, which replace the standard selection and MRA operations. Throughout the paper we focus on learned selection and MRA, but in Appendix \ref{appendix:operators} we outline how to additionally construct sampling and cross-over operators using self-attention.

\section{Meta-Training, Objective \& Tasks}
\label{sec:meta_bbo}

We meta-optimize the weights of the GA attention modules to perform BBO on a family of representative tasks. More specifically, we make use of the previously introduced MetaBBO procedure \citep{les2022} and evolve the LGA parameters to maximize performance on a task distribution of 10 BBOB \citep{hansen2010real} functions. These include functions with different properties, i.e. separability, conditioning, multi-modality (see Table \ref{table:meta_tasks}). At each meta-generation (see Figure \ref{fig:lga}) we start by uniformly sampling a set of BBO tasks and LGA parameters from a meta-evolutionary optimization algorithm. We denote the set of parameters characterizing the LGA by $\theta_i$ for $i=1,...,M$ meta-population members. Afterwards, each LGA is evaluated on all tasks by running an inner loop search. We compute an aggregated meta-performance score to update the meta-EO.
\begin{algorithm}[h]
\caption{MetaBBO Training of Learned Genetic Algorithms}
\begin{algorithmic}[1]
\Inputs{Meta-population $M$, meta-task size $J$, inner loop population size $N$, generations $T$, $\texttt{MetaEO}$ (e.g. OpenAI-ES, \citep{salimans_2017}).}
\State Initialize the meta-search distribution $\mu, \Sigma = \sigma_{meta} I$.
\While{not done}
\State Sample $J$ BBO tasks with $\xi_l, \ \forall l=1,...,J$.
\State Sample LGA candidates: $\theta_i \sim \mathcal{N}(\mu, \Sigma), \ \forall i = 1, ..., M$.
\State Evaluate all $M$ LGA candidates on the same $K$ tasks:
\For{$l = 1, ..., J$}
\For{$i = 1, ..., M$}
\State Initialize search and fitness archives $X^P, \boldsymbol{\sigma}^P, \mathbf{f}^P$.
\For{$t = 0, ..., T-1$}
\State Sample: $\tilde{X}^P, \tilde{\mathbf{f}}^P, \tilde{\boldsymbol\sigma}^P \ \leftarrow \texttt{Sample}(X^P, \mathbf{f}^P, \boldsymbol{\sigma}^P)$.
\State Perform MRA: $\boldsymbol{\sigma}^C \leftarrow \texttt{Adaptation}_{\theta_i}(\tilde{\mathbf{f}}^P, \tilde{\boldsymbol{\sigma}}^{P})$.
\State Mutate: $X^C \leftarrow \texttt{Mutation}(\tilde{X}^P, \boldsymbol{\sigma}^C)$. 
\State Evaluate all children: $f(X^C_{j} | \xi_l), \ \forall j=1,...,N$.
\State Update parent archive:
\State $X^{P'}, \mathbf{f}^{P'}, \sigma^{P'} \leftarrow \texttt{Selection}_{\theta_i}( \mathbf{f}^{C}, \mathbf{f}^{P})$.
\EndFor
\EndFor
\EndFor
\State Collect fitness scores $\left[\left[\left[\{f(x^C_{j,t} | \xi_l)\}_{j=1}^N\right]_{t=1}^T\right]_{l=1}^J | \theta_i \right]_{i=1}^M$.
\State Compute normalized \& aggregated meta-fitness $\{\bar{f}(\theta_i)\}_{i=1}^M$.
\State Update meta-search $\mu', \Sigma' \leftarrow \texttt{MetaEO}(\{\theta_i, \tilde{f}\}_{i=1}^M|\mu, \Sigma)$.
\EndWhile
\end{algorithmic}
\label{algo:metabbo}
\end{algorithm}
The \emph{MetaBBO}-objective is computed based on the collected inner loop fitness scores of each LGA instance, where $\xi_l$ denotes task-specific parameters for $l=1,...,J$ tasks. For each candidate $\theta_i$ we minimize the final performance of the best population member. Afterwards, we $z$-score the task-specific results over meta-population members and compute the median across tasks:

\vspace{-0.5cm}
$$\left[[f(\theta_i | \xi_k)]_{i=1}^M\right]_{k=1}^K = \left[ \left[ \min_j \{f(x_{j,T} | \xi_l)\}_{j=1}^N | \theta_i \right]_{i=1}^M\right]_{l=1}^J$$
\vspace{-0.25cm}
$$\{\bar{f}(\theta_i)\}_{i=1}^M = \texttt{Median}\left[\texttt{Z-Score}\left([f(\theta_i | \xi_k)]_{i=1}^M\right)\right]_{k=1}^K.$$
%
The outer loop optimizes $\theta$ using  OpenAI-ES \citep{salimans_2017} for 750 meta-generations with a meta-population size of $M=512$. We sample $J=256$ tasks and evaluate each sampled LGA on each task independently. Each LGA is unrolled for $T=50$ generations, with $N=16$ population members in each iterations. 
Throughout, we assume that $E=N$, i.e. the number of parents is equal to the number of children. This is not a limitation since the selection operator is capable of replacing multiple parents with the same child (see Section \ref{sec:viz_lga}).
The large-scale parallel meta-evaluation of all $\theta_i$ on all $\xi_j$ tasks is facilitated by the auto-vectorization and device parallelism capabilities provided by the JAX library \citep{jax2018github, evosax2022github} and runs on multiple accelerators. We provide more details and experiments on the meta-training settings and robustness in Appendix \ref{appendix:metabbo} and \ref{appendix:meta_results}.

\section{Experiments}
\label{sec:experiments}

We now turn to an exhaustive experimental evaluation of the MetaBBO optimization procedure and the discovered LGA. We thereby set out to answer the following questions:

\begin{enumerate}
    \item Is it possible to meta-evolve competitive LGAs via MetaBBO using a limited set of meta-training BBO tasks (Section \ref{sec:meta_train})?
    \item Does the resulting LGA outperform GA baselines on unseen BBO problems and different search budgets (Section \ref{sec:meta_test})?
    \item How much can an LGA discovered on a limited set of tasks generalize beyond its meta-training setting (e.g. hyperparameter optimization \& neuroevolution; Sections \ref{sec:hpo_b} \& \ref{sec:neuroevolution})?
\end{enumerate}

\subsection{Meta-Training on BBOB Functions}
\label{sec:meta_train}

\begin{figure}[tb]
\vspace{-0.5cm}
\caption{LGA MetaBBO. Meta-evaluation across meta-generations. Evaluation of LGA on two 10 dim. BBOB (\underline{Top}) and neuroevolution tasks (\underline{Bottom}). Mean \& 1.96 standard error intervals across 3 independent MetaBBO runs.}
\centering
\includegraphics[width=0.475\textwidth]{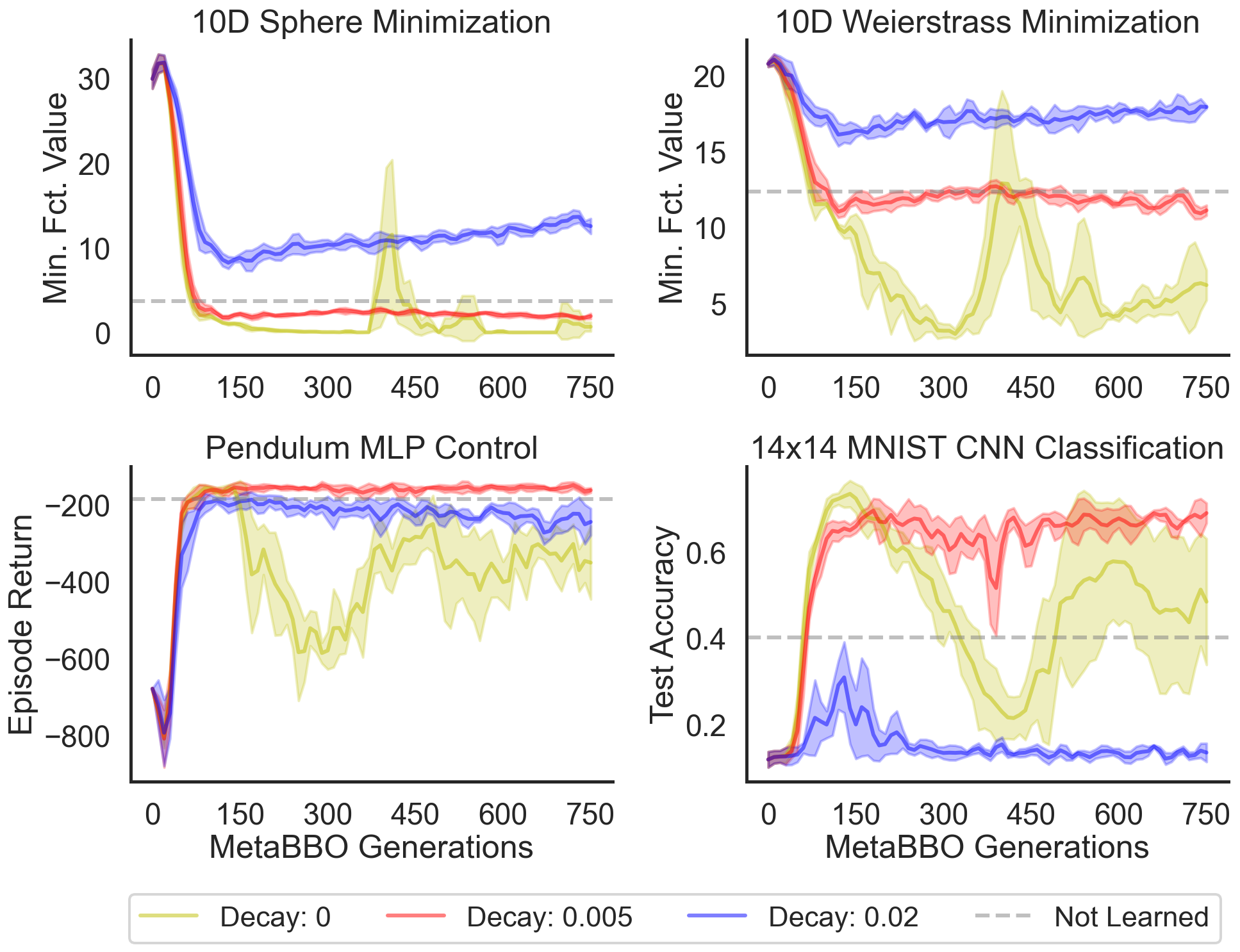}
\label{fig:meta_train}
\vspace{-0.5cm}
\end{figure}

We start by meta-evolving the LGA parameters on a task distribution consisting of 10 BBOB functions with different random optima offsets, evaluation noise and considered problem dimensionality ($D \leq 10$). Throughout meta-training we evaluate the performance of the optimized LGA on several different downstream tasks. These include BBOB functions seen during meta-training, hold-out meta-test BBOB functions and small neuroevolution tasks. In Figure \ref{fig:meta_train} we plot the detailed evaluation curves across meta-training.
The MetaBBO-trained LGA quickly learns how to perform optimization on the low-dimensional BBOB meta-training functions. Interestingly, we find that the MetaBBO procedure can lead to an LGA that overfits to the BBOB tasks on which it was meta-trained. The downstream performance of LGA decreases and becomes unstable for both an unseen Pendulum control task with MLP policy and a downsized 14-by-14 MNIST classification task using a CNN. 
We therefore investigated whether meta-regularization can improve the generalization on such unseen neuroevolution tasks. More specifically, we compared three different meta-mean regulatization coefficients $\lambda \in \{0, 0.005, 0.02\}$, which exponentially decay the meta-mean to zero, $\mu_\lambda' = (1 - \lambda) \mu'$.
We observe that the generalization to the neuroevolution tasks can be improved and stabilized using a properly chosen decay of $0.005$.
In Appendix \ref{appendix:meta_results} we further explore the impact of the meta-task distribution, meta-objective and LGA attention size. The MetaBBO procedure is largely robust to the choice of these settings.
Small attention layers are sufficient for consistently discovering performant LGAs. This comes with the additional advantage of reducing the FLOPs and memory requirements of executing the LGA. The evaluation of a meta-trained LGA is easily feasible on a single core CPU device. 

\subsection{Meta-Testing on BBOB Functions}
\label{sec:meta_test}

\begin{figure}[tb]
\vspace{-0.5cm}
\caption{LGA Generalization. \underline{Left}: BBOB evaluation for different budgets \& spaces. \underline{Right}: Performance across generations. Mean \& 1.96 standard error intervals across 50 runs.}
\centering
\includegraphics[width=0.425\textwidth]{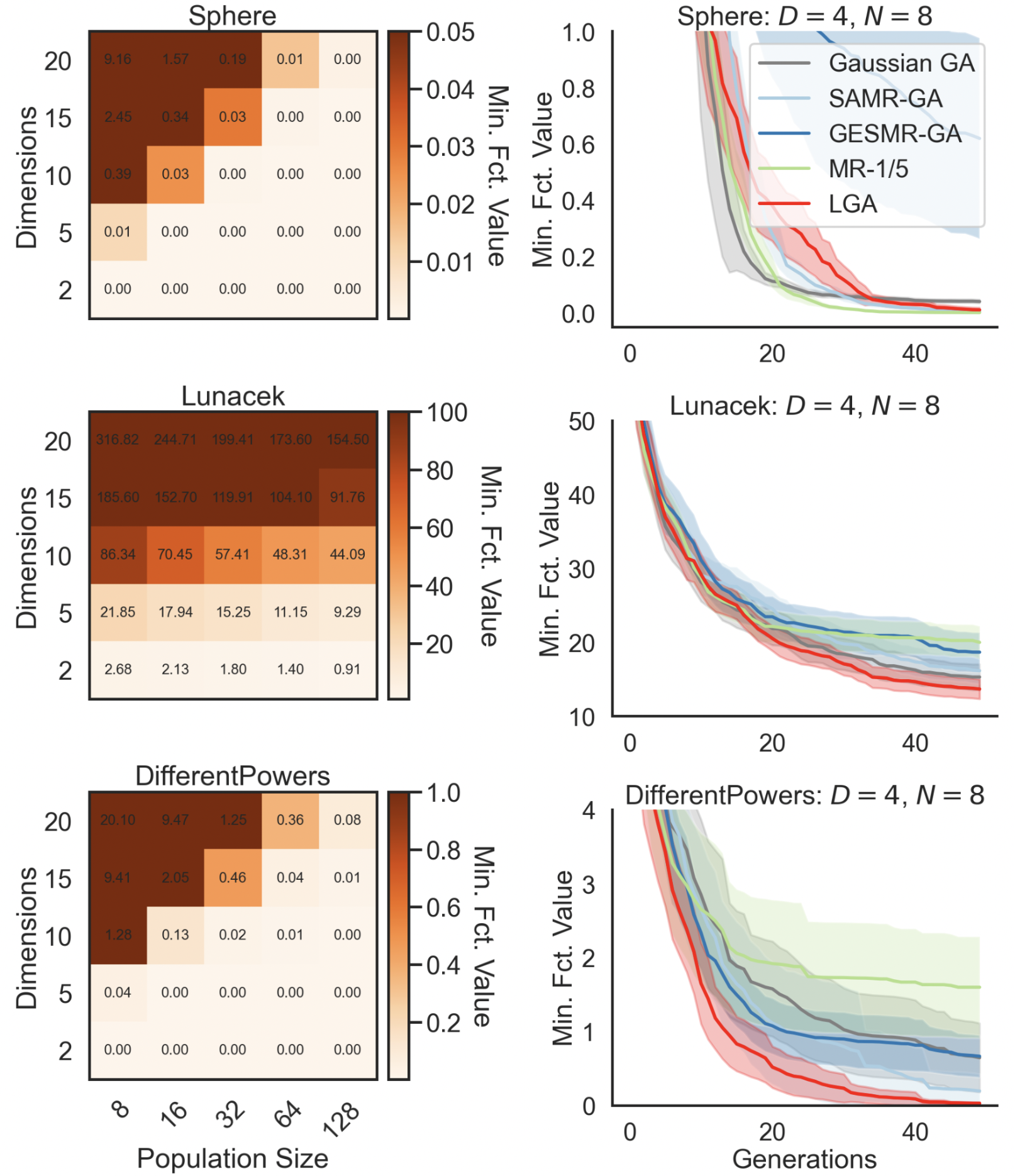}
\label{fig:meta_test_2}
\vspace{-0.5cm}
\end{figure}

Next, we exhaustively evaluate the performance of LGA on the full set of BBOB benchmark functions including test functions unseen during meta-training. We compare against 4 competitive GA baselines: Gaussian GA, MR-1/5 GA, SAMR-GA, GESMR-GA.\footnote{All baselines were tuned using grid searches over the parent archive size and initial mutation rate scale. We report all BBOB evaluation \& tuning settings in Appendix \ref{appendix:bbob_eval_params}.} 
\begin{figure}[tb]
\vspace{-0.5cm}
\caption{Meta-Evaluation on BBOB Tasks. \underline{Left}: Training Functions. \underline{Right}: Hold-Out Functions. Scores are normalized by the Gaussian GA baseline performance. Lower is better. Averaged over 50 independent evaluation runs.}
\centering
\includegraphics[width=0.475\textwidth]{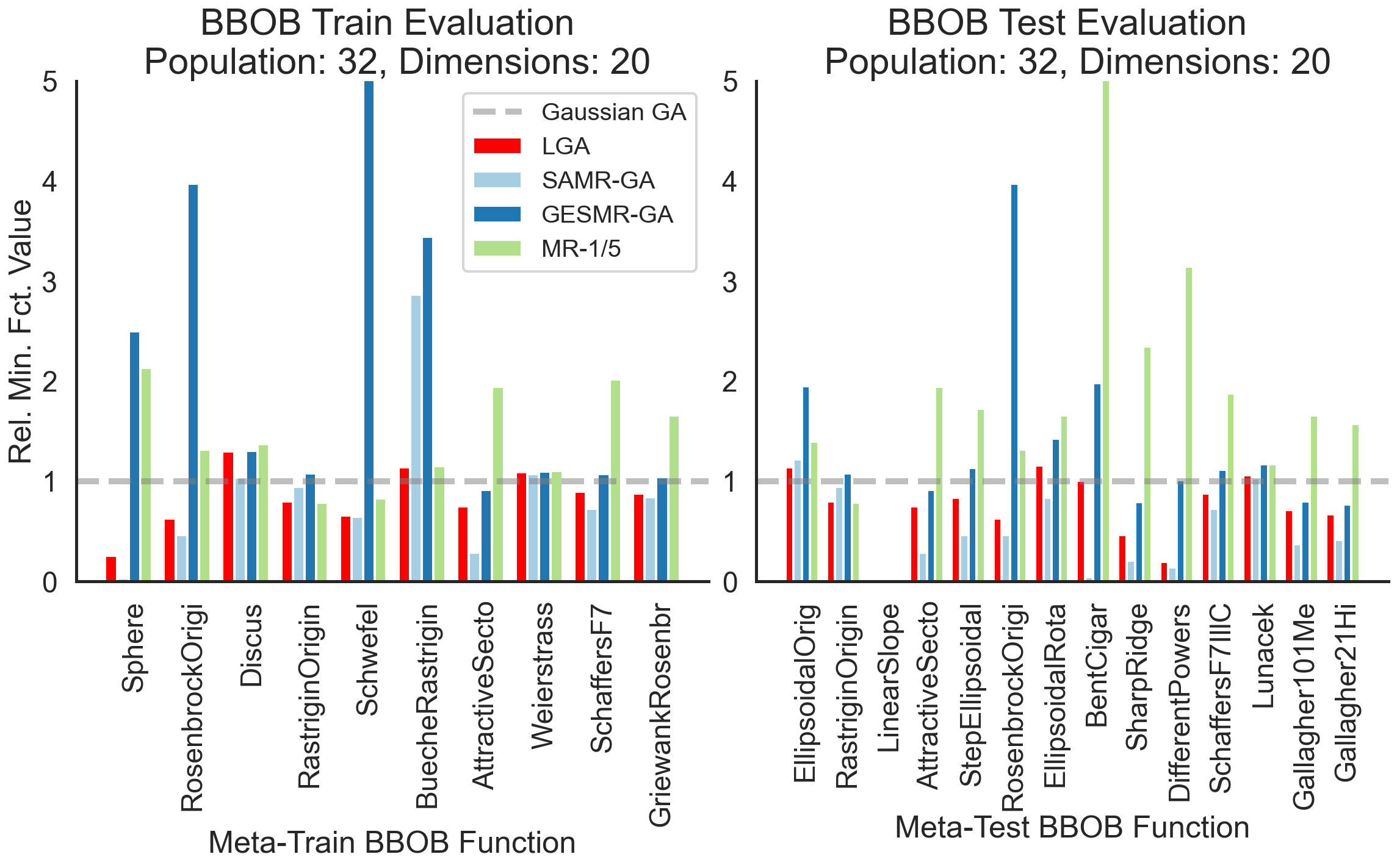}
\label{fig:meta_test_1}
\vspace{-0.5cm}
\end{figure}
We compare the performance on all BBOB functions for a population size of $N=32$, $D=20$ search dimensions and for $T=50$ generations. The best-across generations function value is normalized by the performance of the Gaussian GA. In Figure \ref{fig:meta_test_1} we find that LGA outperforms all baselines on the majority of both BBOB functions seen during meta-training (left) and unseen BBOB functions (right). This holds true for functions with very different characteristics (single/multi-modal, high/low conditioning, separable/not separable), search dimensions and population sizes (Appendix Figure \ref{fig:si_bbob_train} \& \ref{fig:si_bbob_test}). This provides further evidence that LGA does in fact not overfit to the BBO functions seen during meta-training, but instead has discovered a general-purpose GA algorithm.
In Figure \ref{fig:meta_test_2} we further demonstrate that LGA generalizes to different population sizes and problem dimensions. The meta-learned GA achieves lower function values in fewer generations (right) and performs well for different problem settings (left). As the problems become harder with increased dimensionality, LGA can compensate with a larger population size.

\subsection{Meta-Testing on Continuous HPO-B}
\label{sec:hpo_b}

\begin{figure}[tb]
\vspace{-0.75cm}
\caption{LGA Evaluation on HPO-B \citep{arango2021hpo}. \underline{Left}: Small population size ($N=4$). \underline{Right}: Large population size ($N=8$). Mean \& 1.96 standard error intervals across 5 runs.}
\centering
\includegraphics[width=0.475\textwidth]{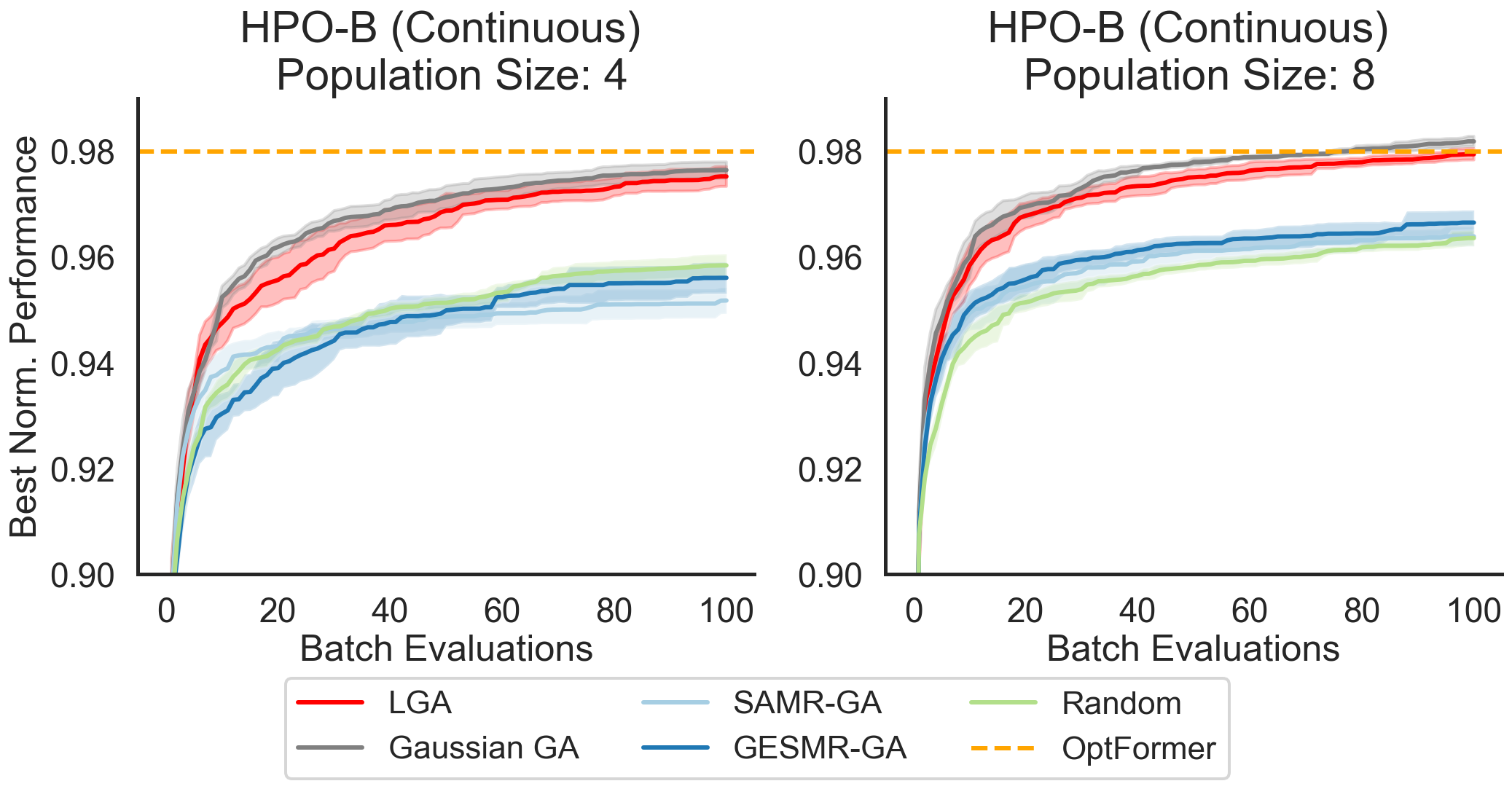}
\label{fig:hpo_b}
\vspace{-0.75cm}
\end{figure}

Next, we test LGA's performance on the HPO-B benchmark \citep{arango2021hpo}. The benchmark considers a vast array of hyperparameter optimization tasks including 16 different model types (SVM, XGBoost, etc.) and their respective search spaces ($D \in \{2, 3, \dots, 16\}$). Each model is evaluated on 2 to 7 different datasets, which leads to a total of 86 hyperparameter search tasks. We consider the continuous HPO-B version, which uses a previously fitted surrogate model.
Note that the LGA has not been trained on such hyperparameter optimization tasks.
Figure \ref{fig:hpo_b} compares the performance of LGA against the GA and a random search baselines. Additionally, we report the reference performance of the recently proposed OptFormer model \citep{chen2022towards} after 105 total evaluations. We find that LGA outperforms the majority of considered GA baselines. Again, this observation holds for two considered population sizes. LGA can also achieve similar performance as OptFormer, which has been trained on a much more diverse task distribution.
This highlights the transfer capabilities and applicability of LGA to new during meta-training unseen optimization domains.

\subsection{Meta-Testing on Neuroevolution Tasks}
\label{sec:neuroevolution}

Until now we have evaluated the performance of the discovered LGA on moderately small search spaces (i.e. $D \leq 20$). But is it also possible to deploy LGA to neuroevolution settings with thousands of search dimensions and arguably very different fitness landscape characteristics? Again, note that LGA has never explicitly been trained to evolve such high-dimensional genomes and that this requires strong transfer of the learned GA operators. 

\begin{figure}[h]
\caption{LGA Evaluation on neuroevolution tasks including continuous control tasks (\underline{top}), visual control (\underline{middle}) \& computer vision (\underline{bottom}) tasks. Mean \& 1.96 standard error intervals across 5 independent runs.}
\centering
\includegraphics[width=0.475\textwidth]{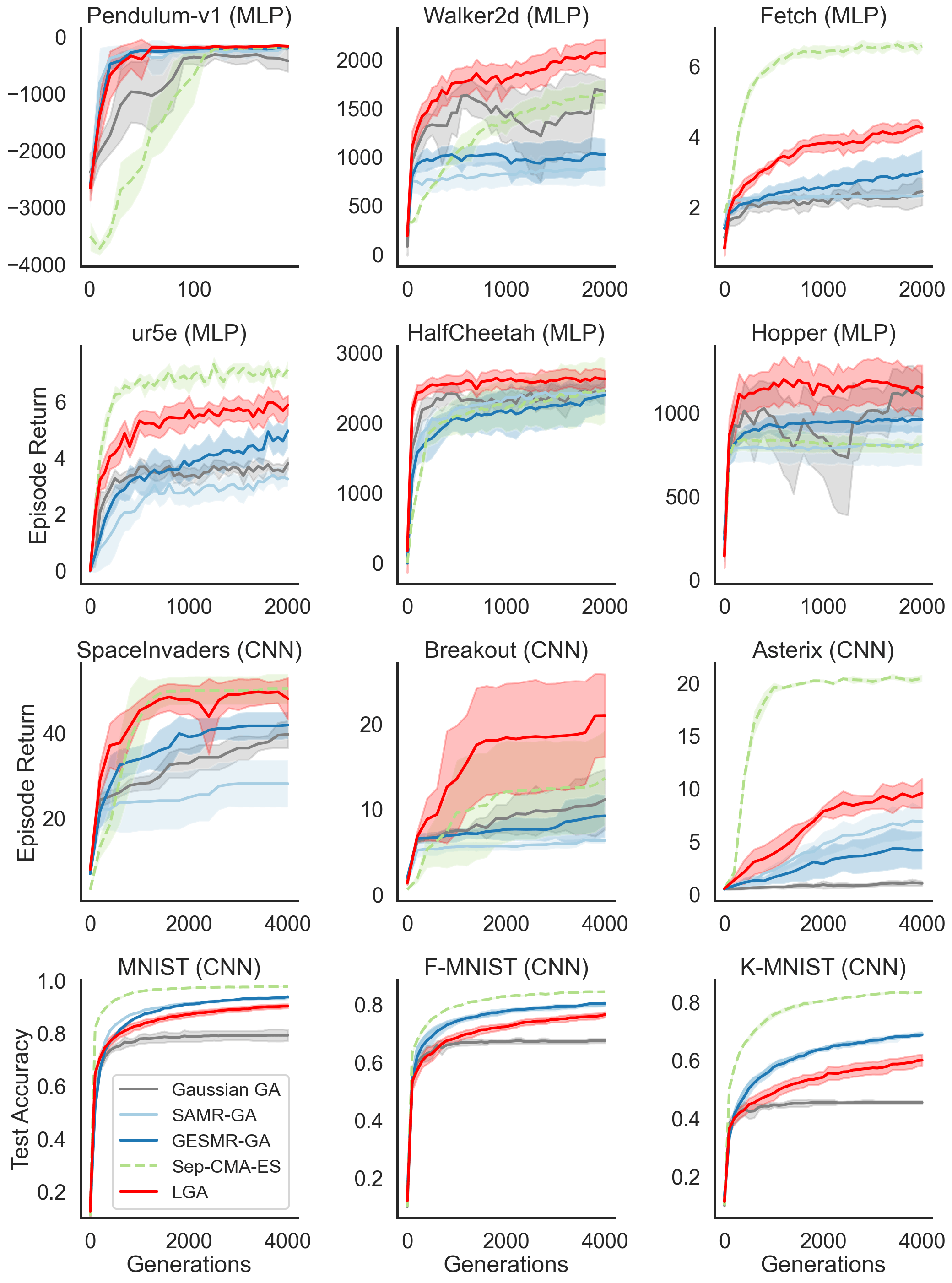}
\label{fig:neuroevo}
\end{figure}

The considered Reinforcement Learning (RL) tasks consist of 4 robotic control tasks (Pendulum-v1 \citep{gymnax2022github} and 5 Brax tasks \citep{freeman_2021}) using MLP policies and three MinAtar visual control tasks (SpaceInvaders, Breakout \& Asterix \citep{young2019minatar}) with CNN-based policies. The MLP genomes consist of less than 1000 weights, while the MinAtar CNNs have ca. 50,000. Hence, the search space is many orders higher than what the LGA has been meta-trained on ($D\leq 10$).
The top three rows of Figure \ref{fig:neuroevo} show that LGA can compete with all tuned baseline GAs on the nine RL tasks with different search spaces, fitness landscapes and evaluation budgets.
Interestingly, the performance gap between LGA and the considered baselines is the biggest for the CNN policies and tends to increase with the number of search dimensions.
Finally, in the final row of Figure \ref{fig:neuroevo} we show that LGA can also successfully be applied to three image classification tasks including MNIST, Fashion-MNIST and K-MNIST classification (28-by-28 grayscale images) with a small CNN (2 convolutional layers, ReLU activation and a linear readout) with 11274 evolvable weights.
The LGA generalizes far beyond the meta-training search horizon($T = 50$ versus 4000) and does not meta-overfit \citep{lange2022learning}.
%
%

\section{Analysis of Discovered LGA}

After having established that the meta-trained LGA is capable of outperforming a set of GA baselines on unseen optimization problems, we now investigate the underlying discovered mechanisms, transfer ability and robustness of the meta-learned GA operators. 


\subsection{Visualization of Learned Genetic Operators}
\label{sec:viz_lga}
\begin{figure}[h]
\caption{LGA's selection operator on a 2-dim Sphere task. \underline{Left}: Sampled selection matrix $S_{:, 1:N} \in \mathbb{R}^{E \times N}$ \underline{Middle}: Matrix indicating whether a child improves the fitness score over the parents. \underline{Right}: Mutation rates for different children. Rows indicate 4 different generations $t \in \{1, 16, 31, 46\}$.}
\centering
\includegraphics[width=0.425\textwidth]{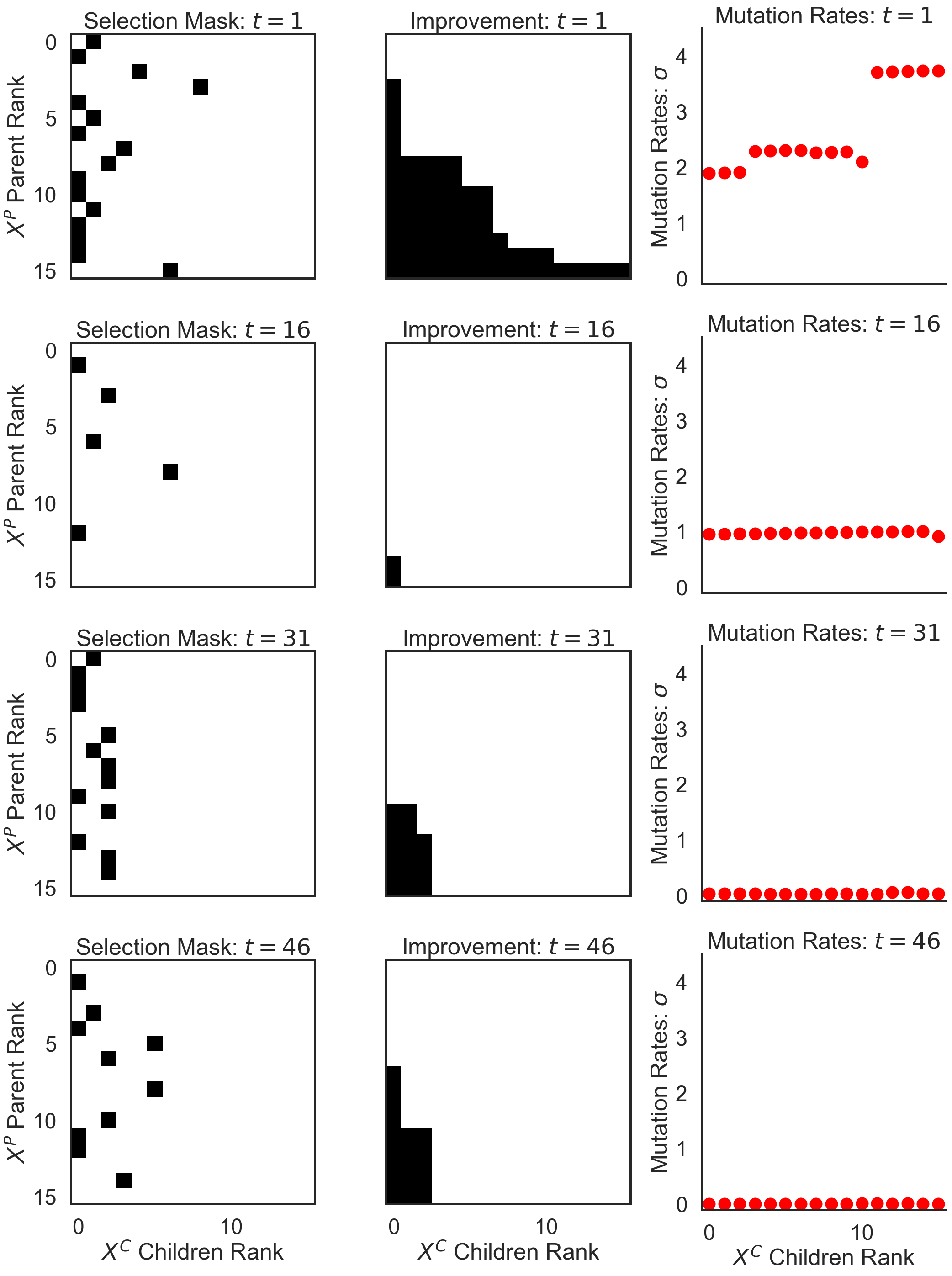}
\label{fig:viz_select}
\end{figure}

What types of mechanisms underlying the black-box genetic operators has LGA discovered? Has it simply re-discovered fitness-based truncation selection or a more complex parent replacement procedure?
In Figure \ref{fig:viz_select} we consider a 2-dim Sphere problem and visualize the selection mask $S_{:, 1:N}$ used to update the parent archive $X^P$. We observe that the selection operator uses children solution to replace parents based on their improvement over the best seen solution. Furthermore, one well-performing child often times replaces more than a single parent. This indirectly implies that the selection operator has meta-learned to dynamically adapt its elite archive size and thereby also the effective sampling distribution. A child that has replaced multiple parents will be sampled (with replacement) more frequently in the next generation. Furthermore, this implies a robustness mechanism: Since children can be stored multiple times in the parent archive, they are less likely to be `forgotten' by the stochastic selection. The mutation rates, on the other hand, are decreased over the course of generations in order to explore closer to the global optimum. Furthermore, we observe a grouping of the MR based on the performance of the parents. Children with bad performing parents tend to exhibit a higher mutation rate.

\vspace{-0.5cm}
\vspace{-0.5cm}

\vspace{-0.25cm}

\subsection{Ablation \& Transfer of Genetic Operators}
\label{sec:ablation}

\begin{figure}[h]
\caption{Visualization of LGA's operator ablations during MetaBBO. Evaluation of LGA on two 10 dim. BBOB (\underline{Top}) and neuroevolution tasks (\underline{Bottom}). We report mean \& 1.96 standard error intervals across 3 independent MetaBBO runs.}
\centering
\includegraphics[width=0.45\textwidth]{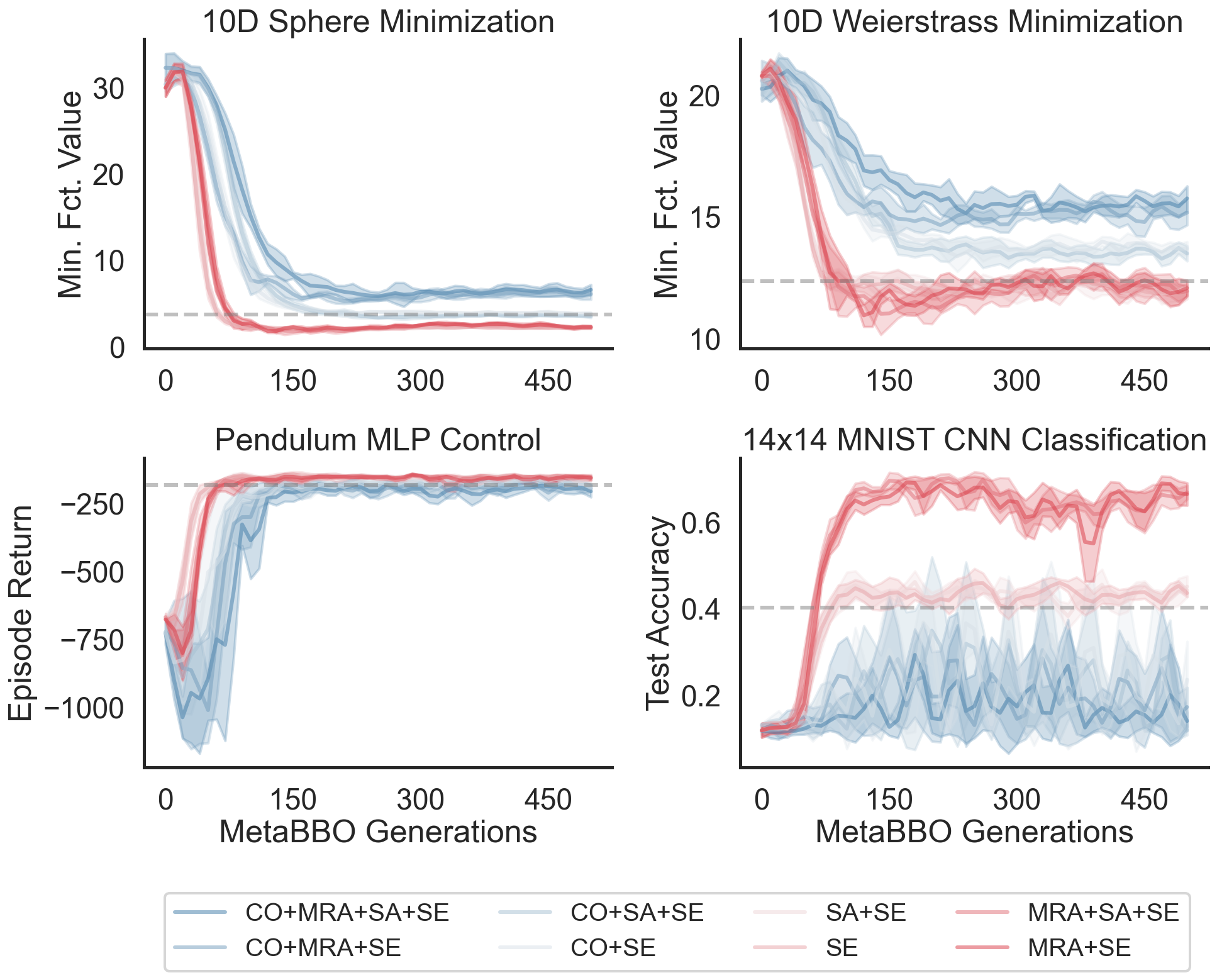}
\label{fig:ablations}
\end{figure}

How much do the different learned components contribute to the overall performance of LGA? Can the learned modules act as drop-in replacements for other genetic algorithms? To answer this question we consider two types of comparative studies:

\begin{enumerate}
    \item \textbf{Operator ablation before MetaBBO discovery}: We meta-train the LGA with a variable amount of learned genetic operators. E.g. we fix the selection operator to white-box truncation selection and only meta-learn mutation rate adaptation. This allows us to quantify the joint contributions and synergies between the different learned ingredients. 
    \item \textbf{Operator transfer after MetaBBO discovery}: After meta-training is completed, we ask whether or not it is possible to substitute the learned operators into other genetic algorithms? This in turn allows us to assess whether the specific learned operator is overfit to the downstream GA computations or whether it can act as a transferable inductive bias for genetic computation.
\end{enumerate}
For the first study we compare meta-training combinations of attention-parametrized selection (SE), mutation rate adaptation (MRA), cross-over (CO) and sampling (SA).\footnote{CO and SA parametrizations are additionally introduced in Appendix \ref{appendix:operators}. Throughout the main text we mainly focused on learned mutation rate adaptation and selection.} For each combination we plot the evaluation performance across meta-generations in Figure \ref{fig:ablations}.
We observe that MRA is crucial for good performance on the neuroevolution tasks. Intuitively, this can be explained by the smaller scale of solution parameters associated with neural network weights. The GA benefits from the ability to flexibly down-regulate its perturbation strengths.
Cross-over, on the other hand, is detrimental for the generalization of LGA to the MNIST CNN neuroevolution task. This behavior can arguably be attributed to the challenge of finding beneficial crossing over pairs for different neural network genomes.
We further observed that learned sampling does not significantly improve the performance of the LGA. We hypothesize that this is due to the indirect effect of the selection mechanism on the sampling of children.
Finally, the overall best performing configuration only meta-learns selection and MRA.

\vspace{-0.5cm}
\vspace{-0.4cm}

\begin{figure}[h]
\caption{Transfer of learned operators to a Gaussian GA. 
Mean \& 1.96 standard error intervals across 5 runs.}
\centering
\includegraphics[width=0.4\textwidth]{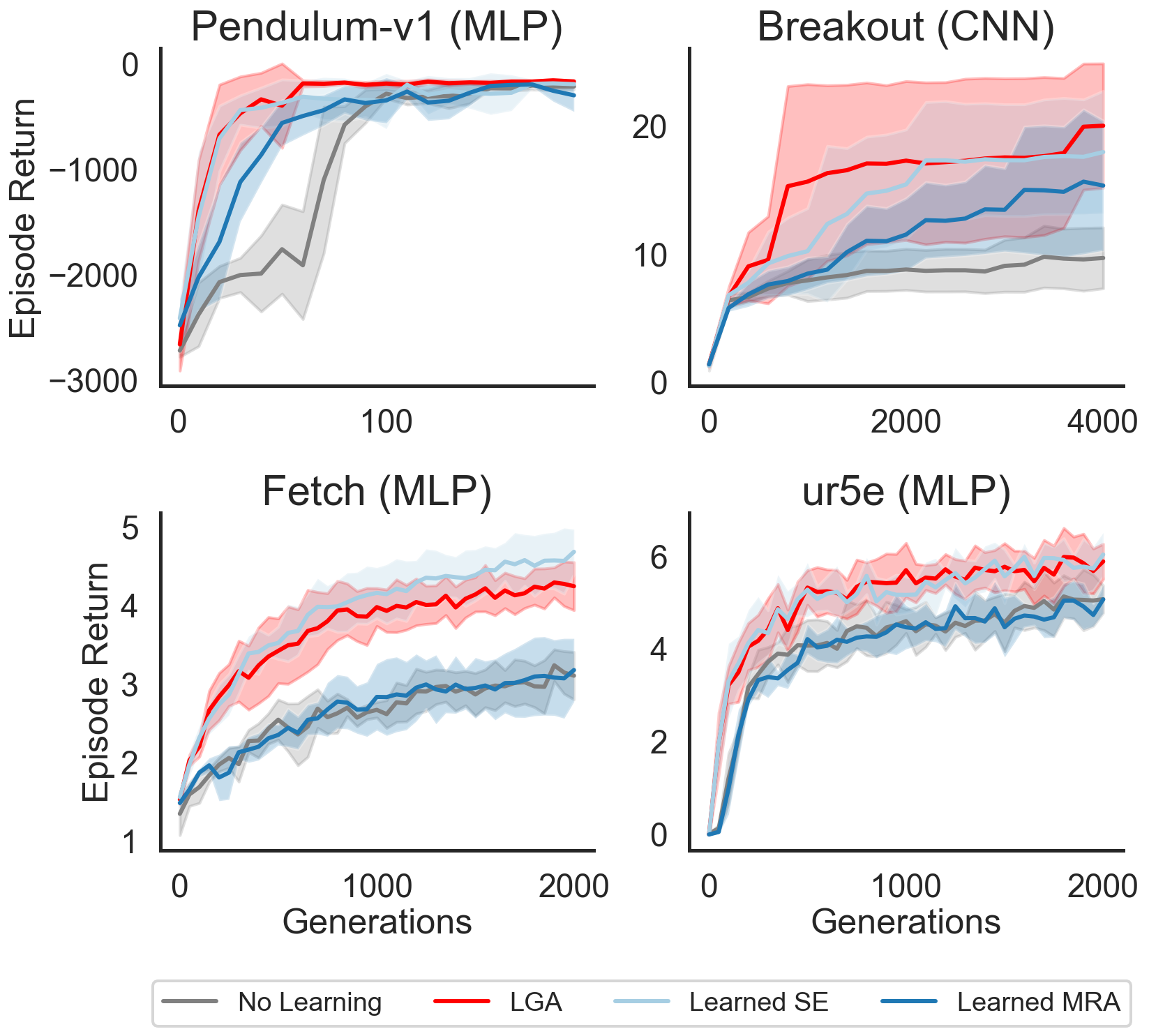}
\label{fig:lga_transfer}
\end{figure}

Next, we considered replacing the truncation selection and fixed mutation rate of the Gaussian GA baseline with the learned selection and MRA operators. In Figure \ref{fig:lga_transfer} we show that this can successfully be accomplished for four neuroevolution tasks. Replacing either the selection or adding learned MRA operator improves the performance of the Gaussian GA. The learned operators can act as drop-in replacements and are transferable inductive biases.

\vspace{-0.5cm}
\vspace{-0.5cm}

\newpage
\subsection{Hyperparameter Robustness of LGA}
\label{sec:robustness}

\begin{figure}[h]
\caption{Hyperparameter Robustness of LGA. Pendulum-v1 performance across elite ratios and initial mutation rates. $\rho = 0$ uses a single parent $E=1$. Results are averaged over 5 independent evaluation runs.}
\centering
\includegraphics[width=0.475\textwidth]{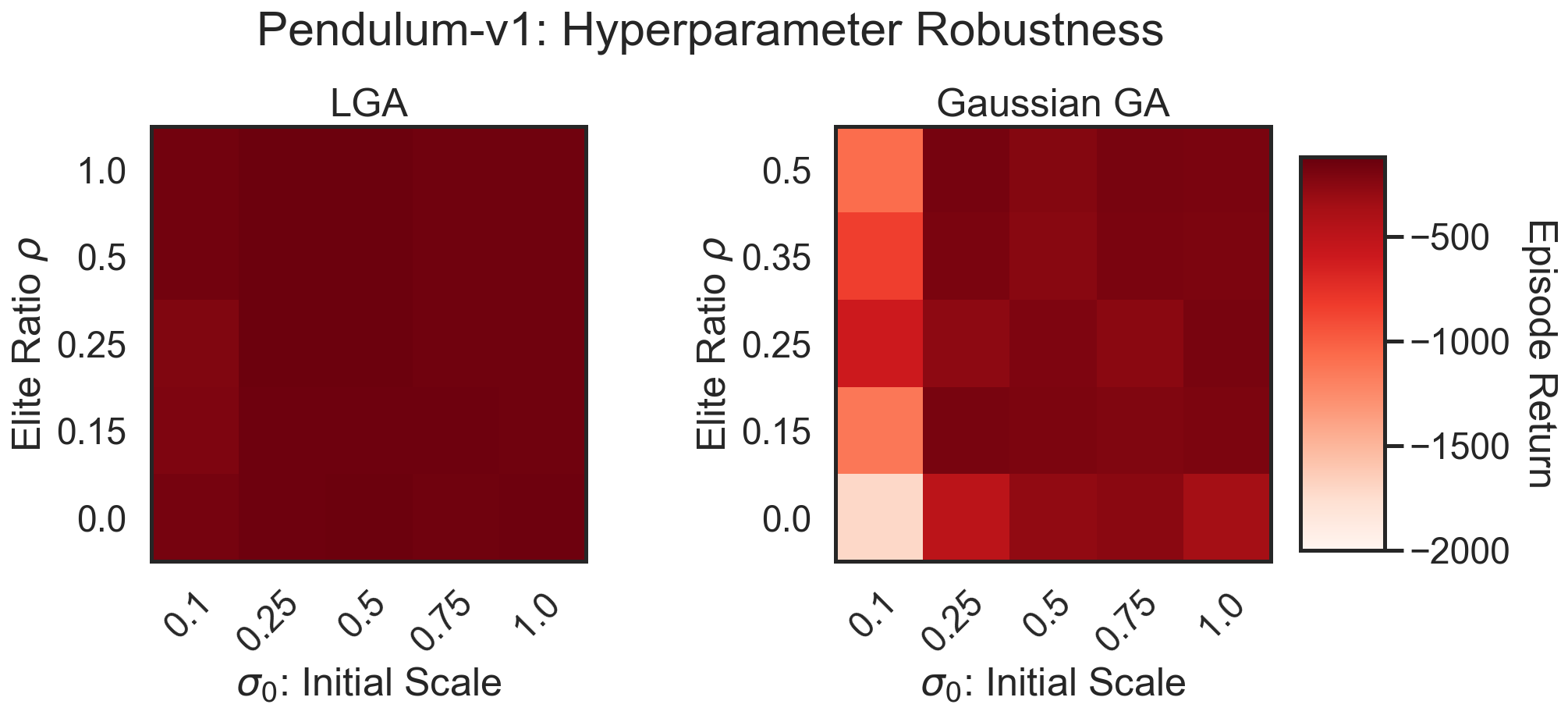}
\label{fig:robustness}
\end{figure}

Finally, we assess the sensitivity of LGA to its remaining hyperparameter choices. More specifically, we compare the performance of LGA and the baseline GAs for various initial mutation rate scales $\sigma_0$ and parent archive sizes $E = \left\lceil \rho \times N \right\rceil$, where $\rho$ denotes the fraction of population members making up the number of parents. Note that while LGA was meta-trained for $\rho=1$, i.e. $E=N$, we find that it is capable of generalizing to many different archive sizes and is robust to the initial scale parameters (Pendulum control task; see Figure \ref{fig:robustness}). In Section \ref{appendix:robustness} we provide the same analysis for all neuroevolution tasks. LGA is far less hyperparameter sensitive than the considered baseline GAs. This highlights the robustness of the LGA induced by the MetaBBO process.

\vspace{-0.5cm}
\vspace{-0.5cm}

\section{Conclusion}
\label{sec:conclusion}
\textbf{Summary}. In this study, we used evolutionary optimization to discover novel genetic algorithms via meta-learning. We leveraged the insight that GA operators perform set operations and parametrized them with novel attention modules that induce the inductive bias of permutation equivariance. Our benchmark results on BBOB, HPO-B and neuroevolution tasks highlight the potential of combining flexible GA parametrization with data-driven meta-evolution.\\
\textbf{Limitations}. We use powerful neural network layers to characterize GA operators. While flexible and interpretable (Section \ref{sec:viz_lga}), the underlying mechanisms do remain partially opaque. Future work needs to be done in order to fully reverse-engineer the discovered operators. In Appendix \ref{appendix:dga} we provide a first set of insights unraveling simple linear relationships between the attention inputs and their outputs. We believe these can guide the design of new `white-box' GAs informed by `grey-box` discovered LGAs.
Furthermore, our analysis highlights the importance of meta-regularization and the potential for the automated design of meta-training curricula.\\
\textbf{Future Work}. We are interested in explicitly regularizing LGAs to maintain diversity in their parent archive. This may provide a bridge to meta-learned quality-diversity methods \citep{cully2017quality}. Furthermore, it may be possible to parametrize a flexible EO algorithm that can interpolate between the exploration of multiple solution candidates as in GA and a single search distribution mode as in traditional evolution strategies. Finally, we believe that better meta-learned GAs can be discovered by simultaneously co-evolving the meta-task distribution and the learned GA.







\newpage

\bibliographystyle{ACM-Reference-Format}
\bibliography{bibliography}

\begin{acks}
This work was funded by DeepMind. We thank Nemanja Rakićević for valuable feedback on this manuscript.
\end{acks}

\appendix


\section{Attention-Based Sampling \& Cross-Over Operators}
\label{appendix:operators}

Next to learned selection and MRA, we additionally experiment with a learned attention-based \texttt{Sample} and \texttt{CrossOver} operator in Section \ref{sec:ablation}. Here we provide their formal definitions.\\
\textbf{Children Sampling Distribution via Self-Attention}. We use the parent fitness $\mathbf{f}^P$ and its transformations $F^P$ together with an age counter $a^P$ and its tanh transformation $\hat{a}^P$ to compute queries, keys and values. Afterwards, the output is projected and normalized into a probability distribution:

\begin{align*}
    \tilde{Q} = [F^P, \hat{a}^P] W_{\tilde{Q}}, K &= [F^P, \hat{a}^P] W_{\tilde{K}} \in \mathbb{R}^{E \times D_K}, \tilde{V} = [F^P, \hat{a}^P] W_{\tilde{V}} \in \mathbb{R}^{E \times 1} \\
    p^P & = \text{softmax}\left(\frac{\tilde{Q} \tilde{K}^T}{\sqrt{D_K}}\right) \tilde{V} \in \mathbb{R}^{E \times 1}\\
    \tilde{X}^P, \tilde{\mathbf{f}}^P, \tilde{\boldsymbol\sigma}^P &= \texttt{Sample}(X^P, \mathbf{f}^P, \boldsymbol{\sigma}^P| p^P).
\end{align*}

\vspace{-0.75cm}
\begin{figure}[h]
\caption{Extra Learned Genetic Operators. \underline{Top}: Self-attention parent sampling probabilities. \underline{Bottom}: Self-attention additive cross-over operator.}
\centering
\includegraphics[width=0.475\textwidth]{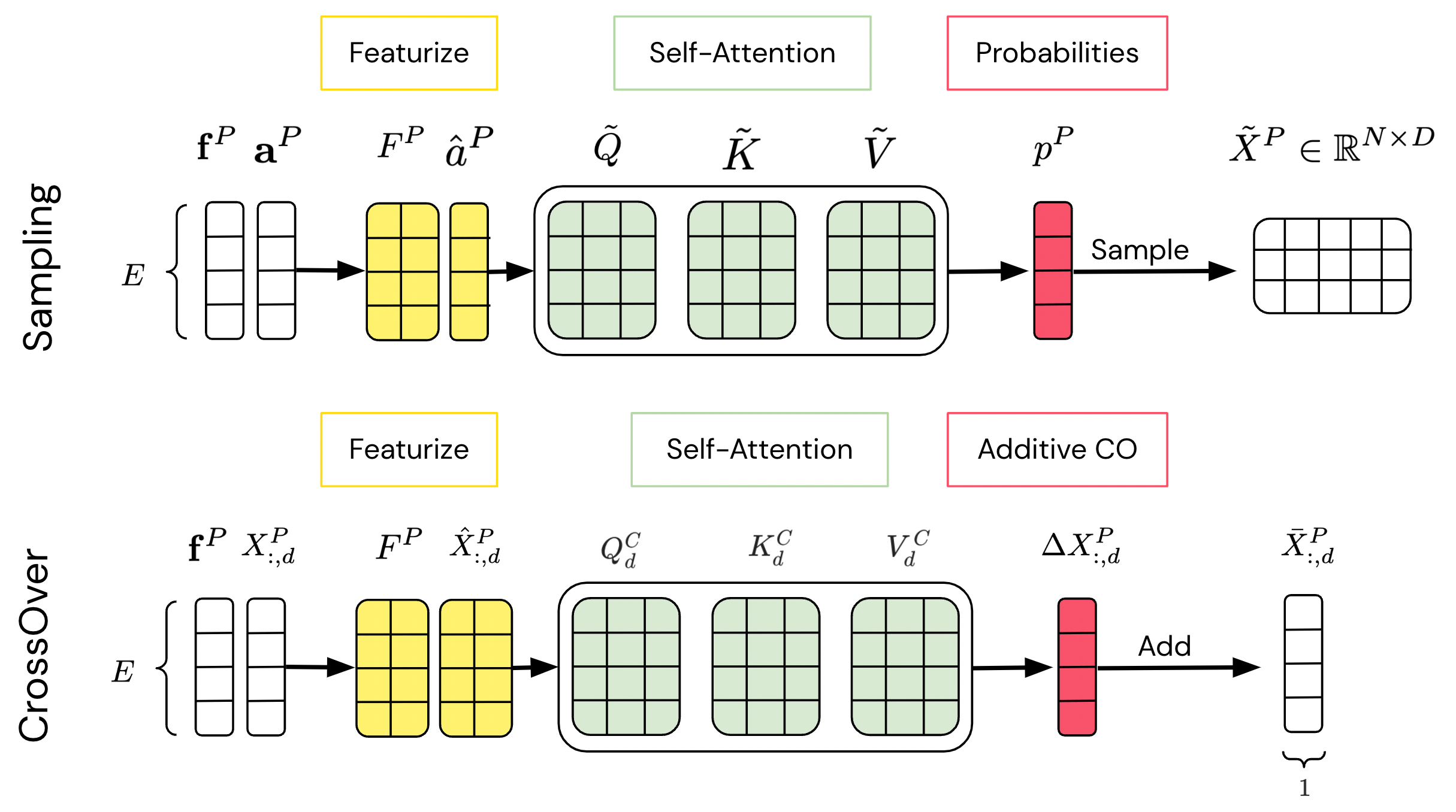}
\label{fig:learned_operators_extra}
\end{figure}
\vspace{-0.5cm}

\textbf{Cross-Over via Self-Attention}. For each dimension $d$ of the parent matrix $X^P_{:, d} \in \mathbb{R}^E$, we first construct a set of normalized features measuring the diversity across all parents (e.g. z-scored feature, normalized distance, etc.) to obtain $\hat{X}^P_{:, d} \in \mathbb{R}^{E \times D_E}$. We then concatenate $[F^{P}, \hat{X}^P_{:, d}] \in \mathbb{R}^{E \times (D_F + D_E)}$, obtain queries, keys and values via embeddings and compute the output of a scaled dot-product self-attention layer:
%
\begin{align*}
    Q^C_{d} &= [F^{P}, \hat{X}^P_{:, d}] W_{Q^C}, K^C_{d} = [F^{P}, \hat{X}^P_{:, d}] W_{K^C},\\
    V^C_{d} &= [F^{P}, \hat{X}^P_{:, d}] W_{V^C}, Z_d = \text{softmax}\left(\frac{Q_{d}^C K_{d}^{C^T}}{\sqrt{D_K}}\right) V_d^C \in \mathbb{R}^{E \times D_K}.
\end{align*}
%
The parameter-specific output $Z_d$ is linearly projected to obtain an additive change to the parents. The dimension cross-over between the parents is then computed as the addition of the cross-over output and the original parent archive dimension vectors $X^P_{:, d}$:

\vspace{-0.5cm}
\begin{align*}
    \Delta X^P_{:, d} &= Z_d W_{\Delta X} \ \text{and} \ \bar{X}^P_{:, d} = X^P_{:, d} + \Delta X^P_{:, d} \in \mathbb{R}^{E \times 1}, \ \forall d=1,...,D.
\end{align*}
\vspace{-0.5cm}

This operation can efficiently be parallelized across dimensions $D$ using for example the auto-vectorization tools provided by JAX.

\section{Hyperparameter Settings}
\label{appendix:settings}

\subsection{MetaBBO Settings for LGA Discovery}
\label{appendix:metabbo}

\begin{table}[H]
\centering
\footnotesize
\begin{tabular}{ |p{1.65cm}|p{2.4cm}|p{2.25cm}|p{1cm}|}
 \hline
 Function & Reference & Property & Tasks\\
 \hline
 Sphere   &  \citet[p.\ 5, ][]{hansen2010real} & Separable (Indep.) & Small\\
 \hdashline
 Rosenbrock   &  \citet[p.\ 40, ][]{hansen2010real}   & Moderate Condition & Medium\\
 Discus & \citet[p.\ 55, ][]{hansen2010real} & High Condition & Medium\\
 Rastrigin   &  \citet[p.\ 75, ][]{hansen2010real}   & Multi-Modal (Local) & Medium\\
 Schwefel & \citet[p.\ 100, ][]{hansen2010real} & Multi-Modal (Global) & Medium\\
 \hdashline
 BuecheRastrigin & \citet[p.\ 20, ][]{hansen2010real} & Separable (Indep.) & Large \\
 AttractiveSector & \citet[p.\ 30, ][]{hansen2010real} & Moderate Condition & Large\\
 Weierstrass & \citet[p.\ 80, ][]{hansen2010real} & Multi-Modal (Global) & Large\\
 SchaffersF7 & \citet[p.\ 85, ][]{hansen2010real} & Multi-Modal (Global) & Large\\
 GriewankRosen & \citet[p.\ 95, ][]{hansen2010real} & Multi-Modal (Global) & Large\\
 \hline
\end{tabular}
\caption{Meta-BBO BBOB-Based Task Families.}
\label{table:meta_tasks}
\end{table}
\vspace{-0.75cm}

We share the task parameters, initialization and fixed randomness for all meta-members $\theta_i$. Fixed stochasticity \& task-based normalization enhances stable meta-optimization \citep[`Pegasus-trick',][]{ng2000pegasus}.

\begin{table}[H]
\centering
\footnotesize
\begin{tabular}{ |p{1.8cm}|p{1.8cm}|p{1.8cm}|p{1.8cm}|}
 \hline
 \textbf{Parameter} & \textbf{Value} & \textbf{Parameter} & \textbf{Value}\\
 \hline \hline
 MetaEO   & OpenAI-ES \citep{salimans_2017}  & $M$: Meta-Pop. & 512\\
 $\alpha_0$, Decay, Final   & $\{0.01, 0.999, 0.001\}$  & $J$: Meta-Tasks & 256\\
 $\sigma_0$, Decay, Final   & $\{0.1, 0.999, 0.001\}$  & Centered ranks & \checkmark\\
 Meta-Objective   & \ref{eq:minNfinalT}  & $D_K$, Att. Heads & $\{16, 2\}$\\
 Inner loop $D$   & $\sim [2, 10]$  & Inner loop $T$ & 50\\
 Inner loop $\sigma_0$ & $\sim [0.01, 0.5]$ & Inner loop $N$ & 16\\
 Offset $x^\star - c$ & $\sim [-5, 5]$ & Inner loop noise & \citet{hansen2009real_noise}\\
 \hline
\end{tabular}
\caption{Meta-BBO Hyperparameters (Figure \ref{fig:meta_train}).}
\label{table:meta_train}
\end{table}
\vspace{-1.0cm}

\subsection{BBOB Evaluation Settings}
\label{appendix:bbob_eval_params}
\begin{table}[H]
\centering
\footnotesize
\begin{tabular}{ |p{1.8cm}|p{1.8cm}|p{1.8cm}|p{1.8cm}|}
 \hline
 \textbf{Parameter} & \textbf{Value} & \textbf{Parameter} & \textbf{Value}\\
 \hline \hline
 Population $N$  & 32  & Dimensions $D$ & 20\\
 Generations $T$  & 50  & $X^P$ Initialization & $\sim [-5, 5]$\\
 \hline
\end{tabular}
\caption{BBOB Hyperparameters (Figures \ref{fig:meta_test_1}, \ref{fig:meta_test_2}).}
\label{table:bbob}
\end{table}
\vspace{-0.5cm}
We tuned the elite ratio $\rho \in \{0.0, 0.15, 0.25, 0.35, 0.5, 1.0\}$ and initial $\sigma_0 \in \{0.1, 0.25, 0.5, 0.75, 1.0\}$ of all baselines \& LGA via a grid sweep.

\subsection{HPO-B (Continuous) Evaluation Settings}
\label{appendix:hpob_eval_params}

Figure \ref{fig:hpo_b} is generated for two different population sizes $N \in \{4, 8\}$ and $T=100$. The GA archives are initialized at $0.5$ and we follow the evaluation protocol outlined in the benchmark repository. We tuned the elite ratio $\rho \in \{0.0, 0.15, 0.25, 0.35, 0.5, 1.0\}$ and initial $\sigma_0 \in \{0.1, 0.25, 0.5, 0.75, 1.0\}$ of all baselines \& LGA via a grid sweep.

\subsection{Neuroevolution Evaluation Settings}

\begin{table}[H]
\centering
\footnotesize
\begin{tabular}{ |p{1.8cm}|p{1.8cm}|p{1.8cm}|p{1.8cm}|}
 \hline
 \textbf{Parameter} & \textbf{Value} & \textbf{Parameter} & \textbf{Value}\\
 \hline \hline
 MNIST $N$   &  128  & MNIST $T$ & 4000\\
 MNIST CNN L  &  8 [5, 5] \& 16 [3,3] & MNIST Batchsize & 1024\\
 \hline \hline
 Brax $N$   &  256  & Brax $T$ & 2000\\
 Norm obs   &  \checkmark  & $X^P$ Initialization & 0\\
 Episode steps  &  500  & MC Evaluations & 8\\
 Brax MLP L & 0 - Only readout & Brax Activation & Tanh \\ 
 \hline \hline
 MinAtar $N$   &  256  & MinAtar $T$ & 4000\\
 MinAtar CNN L   &  16 [3, 3] Filters  & MinAtar MLP L & 1 ReLU 32 units\\
 Episode steps  &  500  & MC Evaluations & 16\\
 \hline
\end{tabular}
\caption{Neurevolution Hyperparameters (Figure \ref{fig:neuroevo}, \ref{fig:lga_transfer}, \ref{fig:robustness}).}
\label{table:neuroevolution}
\end{table}
\vspace{-0.5cm}

MNIST sweep: $\rho \in \{0.0, 0.25, 0.5, 1.0\}$ and $\sigma_0 \in \{0.01, 0.025, 0.05\}$.\\
Brax sweep: $\rho \in \{0.0, 0.25, 0.5, 1.0\}$ and $\sigma_0 \in \{0.025, 0.05, 0.1\}$.\\
MinAtar sweep: $\rho \in \{0.0, 0.25, 0.5, 1.0\}$ and $\sigma_0 \in \{0.05, 0.075, 0.1\}$.\\

\newpage
\section{Additional MetaBBO Results}
\label{appendix:meta_results}

\subsection{Attention Feature Dimension \& Heads}

\vspace{-0.5cm}
\begin{figure}[h]
\caption{LGA MetaBBO - Different model sizes. We report mean/1.96 ste intervals across 3 independent runs.}
\centering
\includegraphics[width=0.375\textwidth]{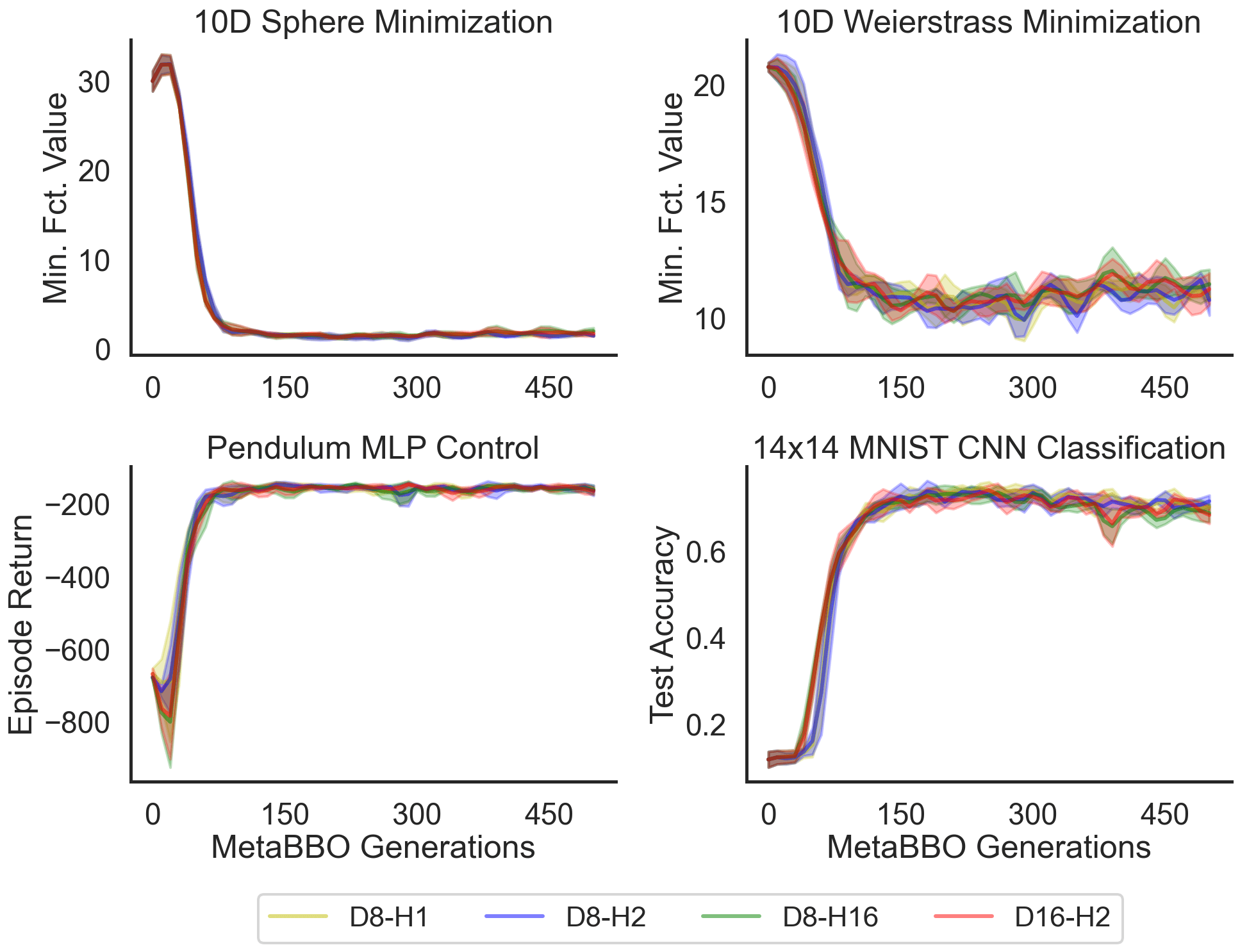}
\label{fig:si_meta_train_model}
\end{figure}
\vspace{-0.5cm}

\subsection{Comparison of Meta-Training Tasks}

\vspace{-0.5cm}
\begin{figure}[h]
\caption{LGA MetaBBO - Different meta-task distributions. We report mean/1.96 ste intervals across 3 independent runs.}
\centering
\includegraphics[width=0.375\textwidth]{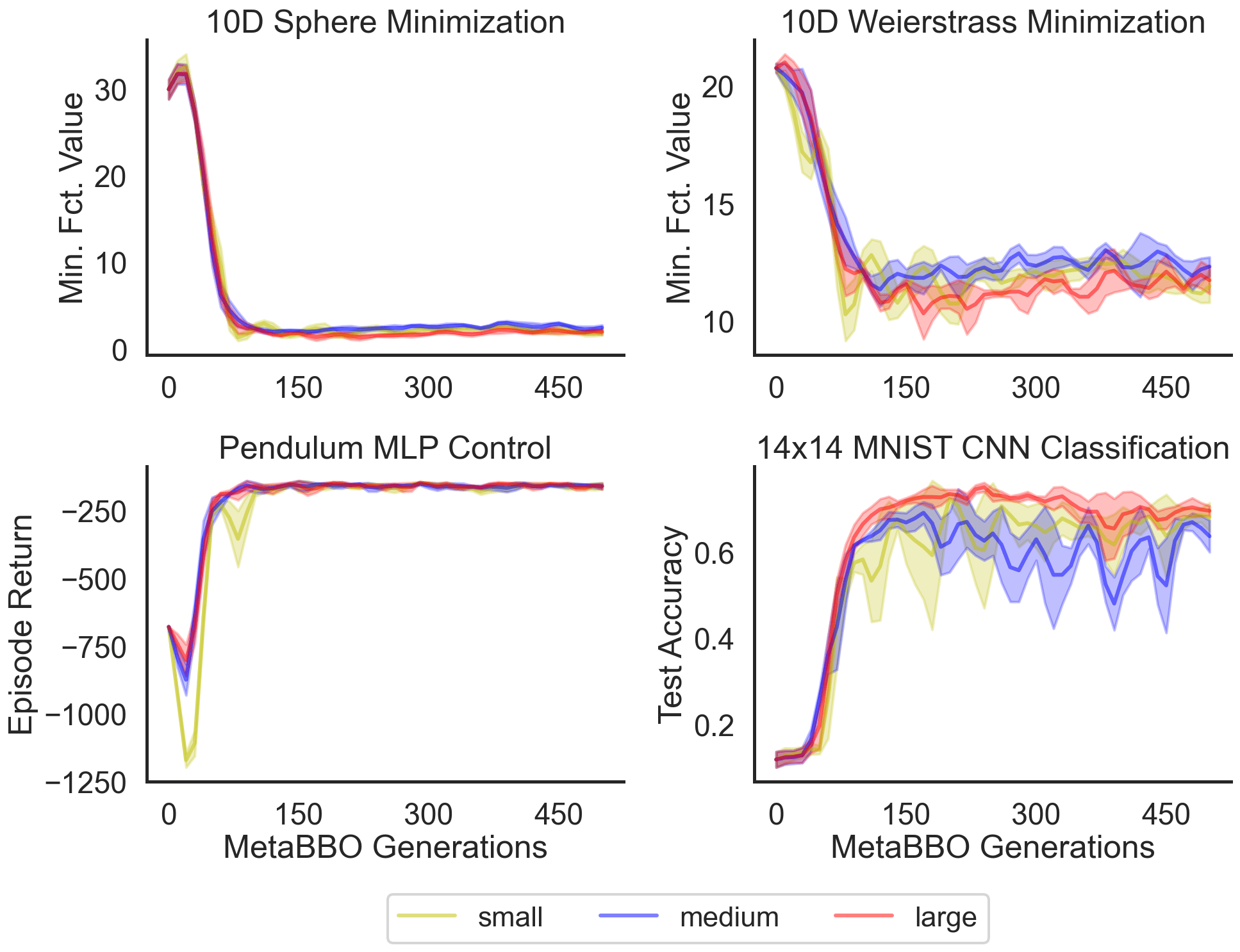}
\label{fig:si_meta_train_tasks}
\end{figure}
\vspace{-0.5cm}

\subsection{Comparison of Meta-Optimizer}
\vspace{-0.5cm}
\begin{figure}[h]
\caption{LGA MetaBBO - Different meta-EO. We report mean/1.96 ste intervals across 3 independent runs.}
\centering
\includegraphics[width=0.375\textwidth]{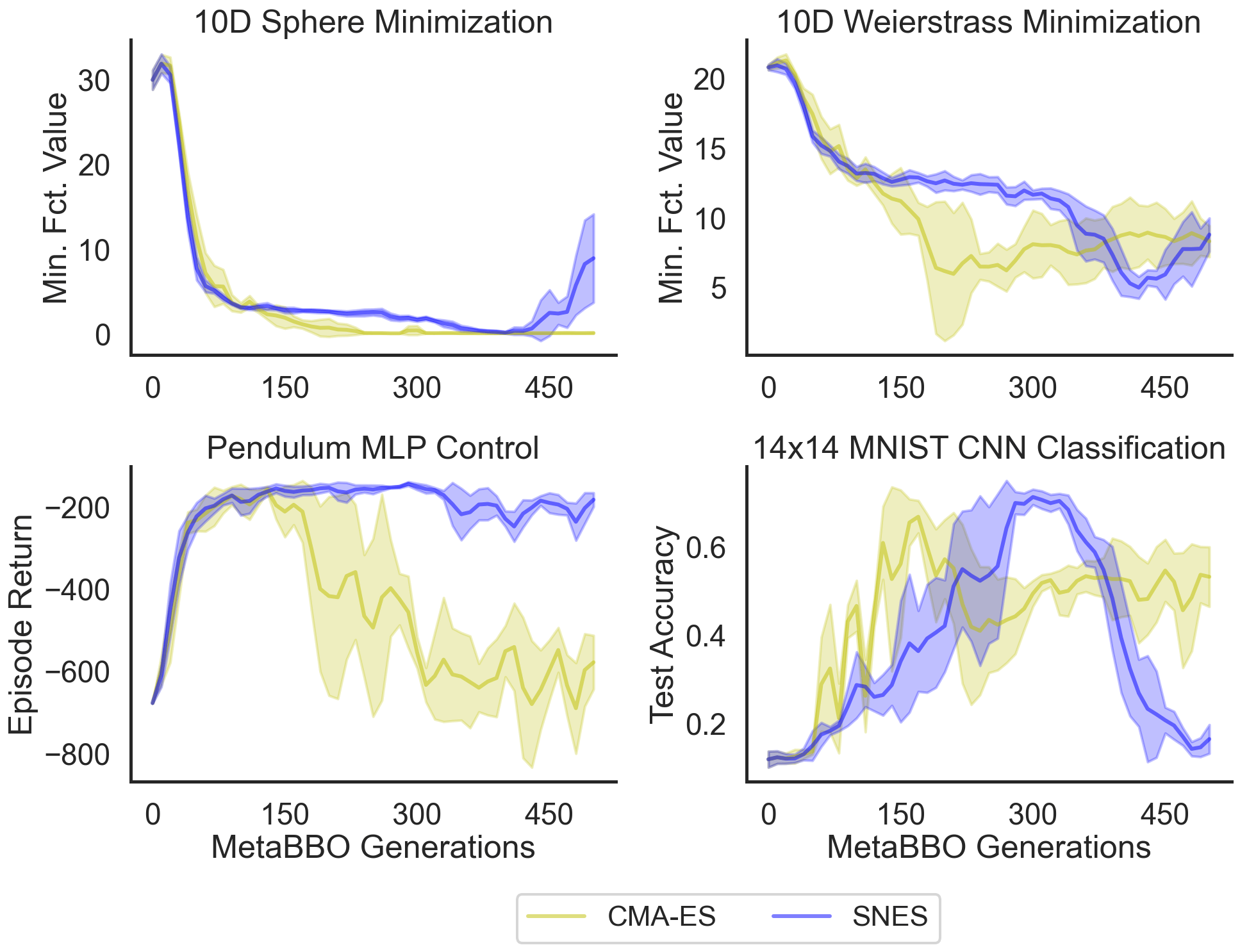}
\label{fig:si_meta_train_tasks}
\end{figure}
\vspace{-0.5cm}

\subsection{Comparison of Meta-Objective Functions}

We also compared differernt meta-objectives, which construct the meta-fitness in different ways:\\
\textbf{\ref{eq:minNminT}}: Minimizes over both the population evaluations with a generation and across all generations.\\
\textbf{\ref{eq:minNfinalT}}: Minimizes over all population evaluations evaluated during the final generation.\\
\textbf{\ref{eq:meanNminT}}: Computes the mean performance across members within a generation \& minimizes this score across generations.\\
\textbf{\ref{eq:meanNfinalT}}: Computes the mean performance across all members within the final generation.
\begin{equation}
    \left[[f(\theta_i | \xi_k)]_{i=1}^M\right]_{k=1}^K = \left[ \left[ \min_t \left[ \min_j \{f(x_{j,t} | \xi_l)\}_{j=1}^N\right]_{t=1}^T | \theta_i \right]_{i=1}^M\right]_{l=1}^J \label{eq:minNminT} \tag{minN-minT}
\end{equation}

\begin{equation}
\left[[f(\theta_i | \xi_k)]_{i=1}^M\right]_{k=1}^K = \left[ \left[ \min_j \{f(x_{j,T} | \xi_l)\}_{j=1}^N | \theta_i \right]_{i=1}^M\right]_{l=1}^J \label{eq:minNfinalT} \tag{minN-finalT}
\end{equation}
\begin{equation}
\left[[f(\theta_i | \xi_k)]_{i=1}^M\right]_{k=1}^K = \left[ \left[ \min_t \left[ \frac{1}{N} \sum_{j=1}^N f(x_{j,T} | \xi_l) | \theta_i\right]_{t=1}^T | \theta_i \right]_{i=1}^M\right]_{l=1}^J \label{eq:meanNminT} \tag{meanN-minT}
\end{equation}
\begin{equation}\left[[f(\theta_i | \xi_k)]_{i=1}^M\right]_{k=1}^K = \left[ \left[ \frac{1}{N} \sum_{j=1}^N f(x_{j,T} | \xi_l) | \theta_i \right]_{i=1}^M\right]_{l=1}^J \label{eq:meanNfinalT} \tag{meanN-finalT}
\end{equation}
\textbf{Interpretation of MetaBBO Comparitive Studies.} The LGA discovery process is largely robust to the considered MetaBBO specifications. Small attention module parametrizations (single head and $D_k = 8$) is sufficient to learn powerful GA operators. Furthermore, a small task distributions (single Sphere function with random offsets/noise) already lead to strong performance on BBOB tasks. For MNIST classification more function diversity is required. The choice of the meta-optimizer and objective are more important. OpenAI-ES (Figure \ref{fig:meta_train}) and \ref{eq:minNfinalT} provide the best performance. 

\begin{figure}[h]
\caption{LGA MetaBBO - Different meta-objectives. We report mean/std intervals across 3 independent runs.}
\centering
\includegraphics[width=0.375\textwidth]{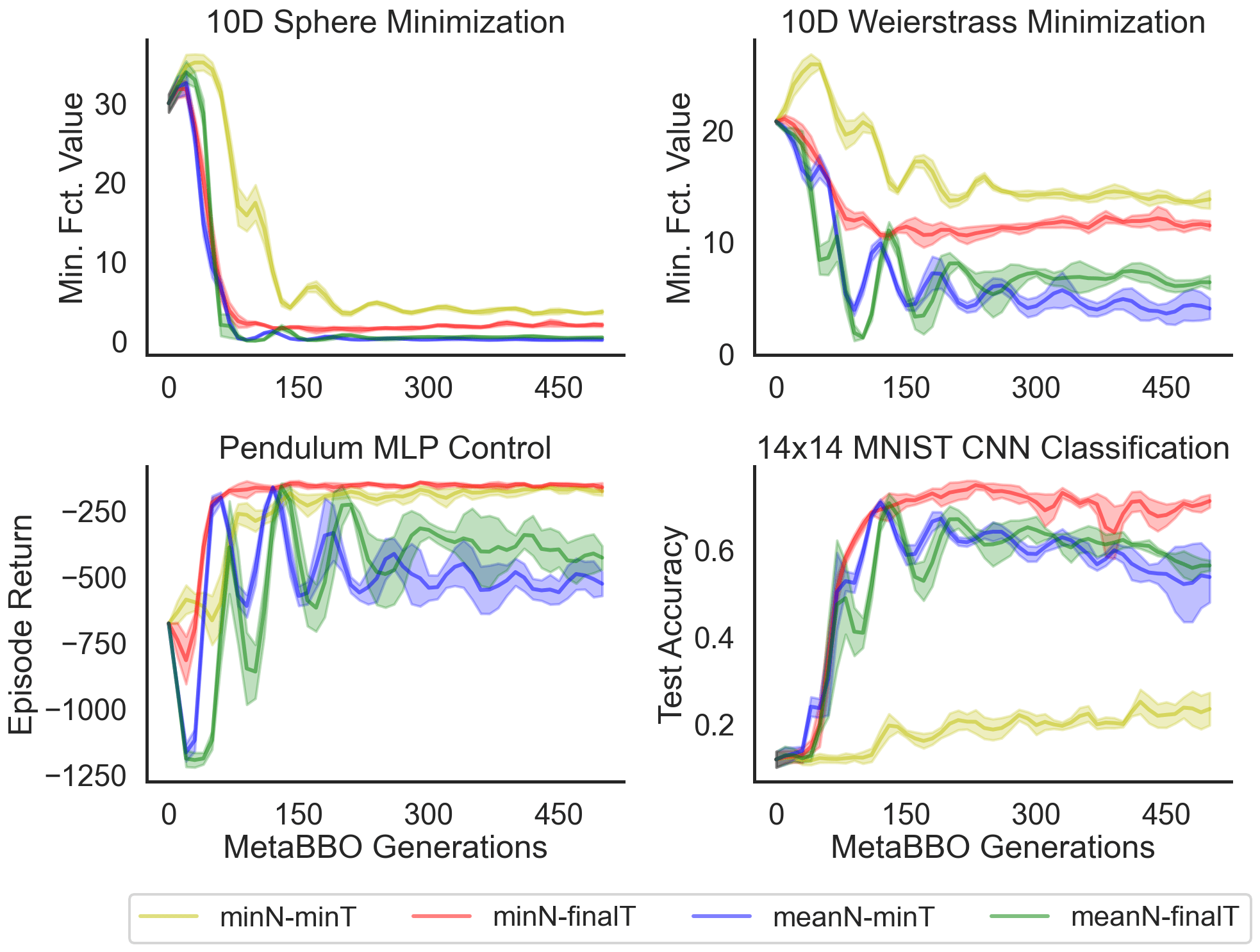}
\label{fig:si_meta_train_objective}
\end{figure}

\newpage
\section{Additional Evaluation Results}
\label{appendix:eval_results}

\subsection{Detailed BBOB Results}
\label{appendix:bbob}

\begin{figure}[h]
\caption{Detailed BBOB Train Function Evaluation.}
\centering
\includegraphics[width=0.475\textwidth]{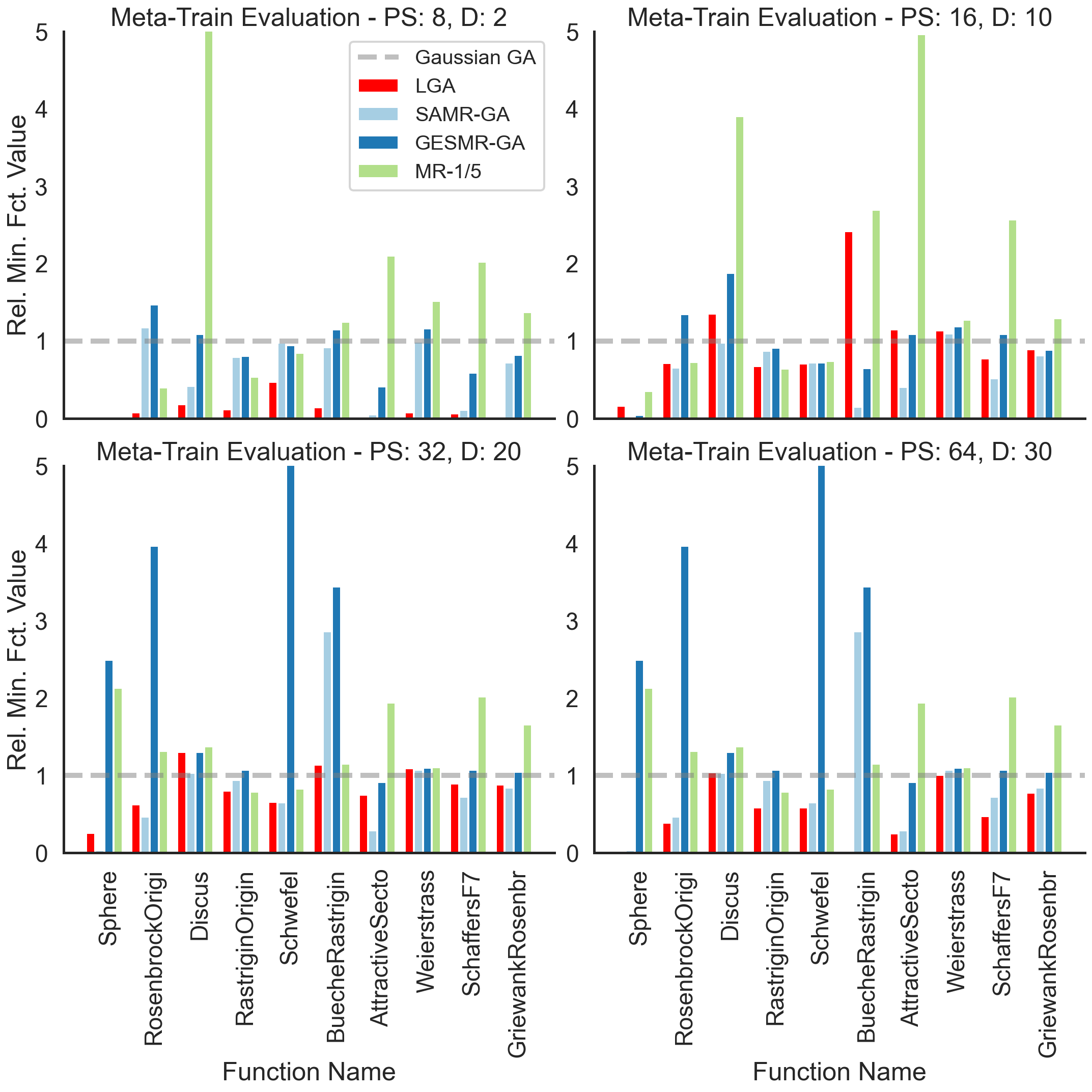}
\label{fig:si_bbob_train}
\end{figure}

\begin{figure}[h]
\caption{Detailed BBOB Test Function Evaluation.}
\centering
\includegraphics[width=0.475\textwidth]{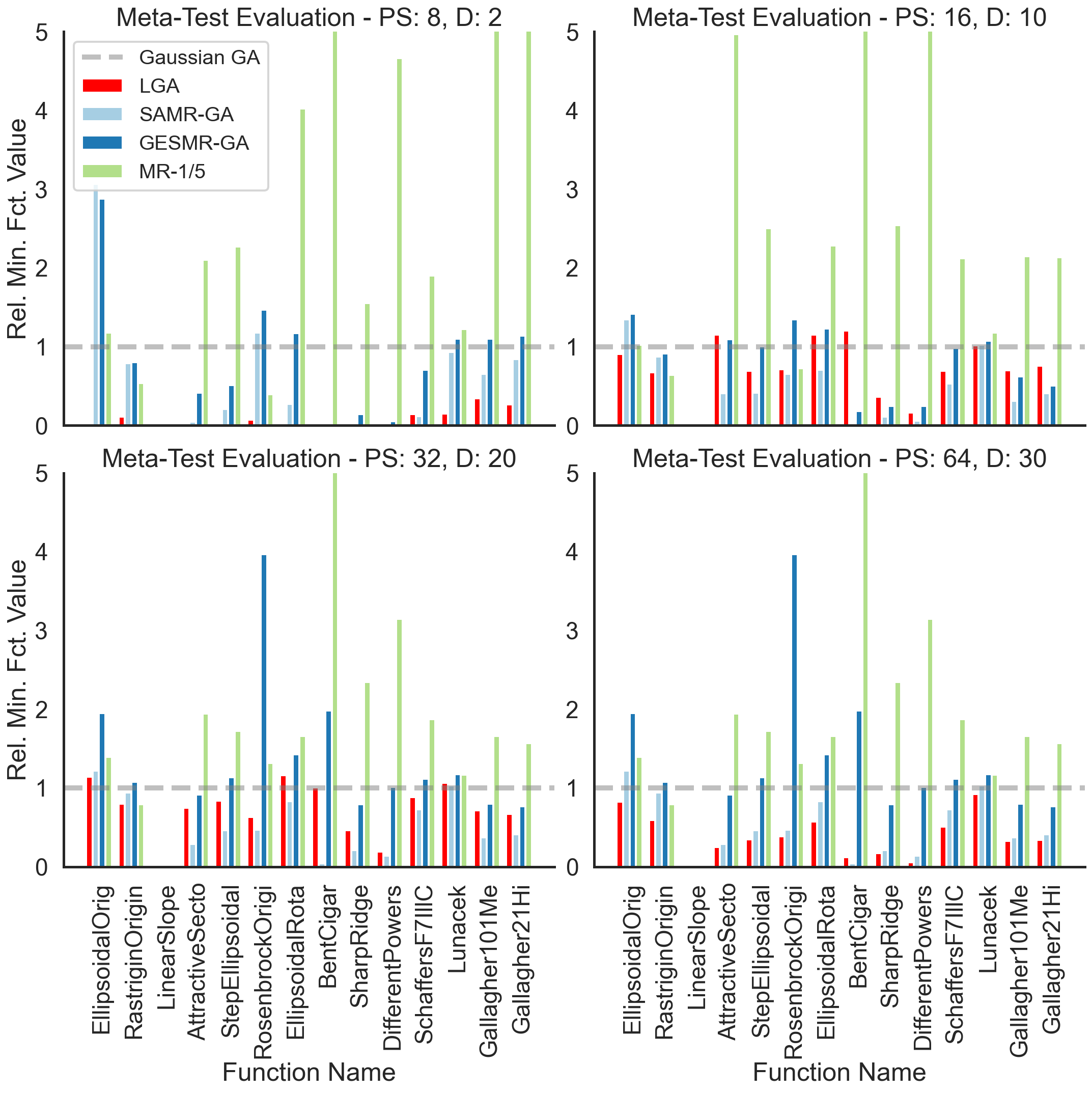}
\label{fig:si_bbob_test}
\end{figure}

\subsection{Neuroevolution Parameter Robustness}
\label{appendix:robustness}

\begin{figure}[h]
\caption{Hyperparameter Robustness - MNIST Task.}
\centering
\includegraphics[width=0.4\textwidth]{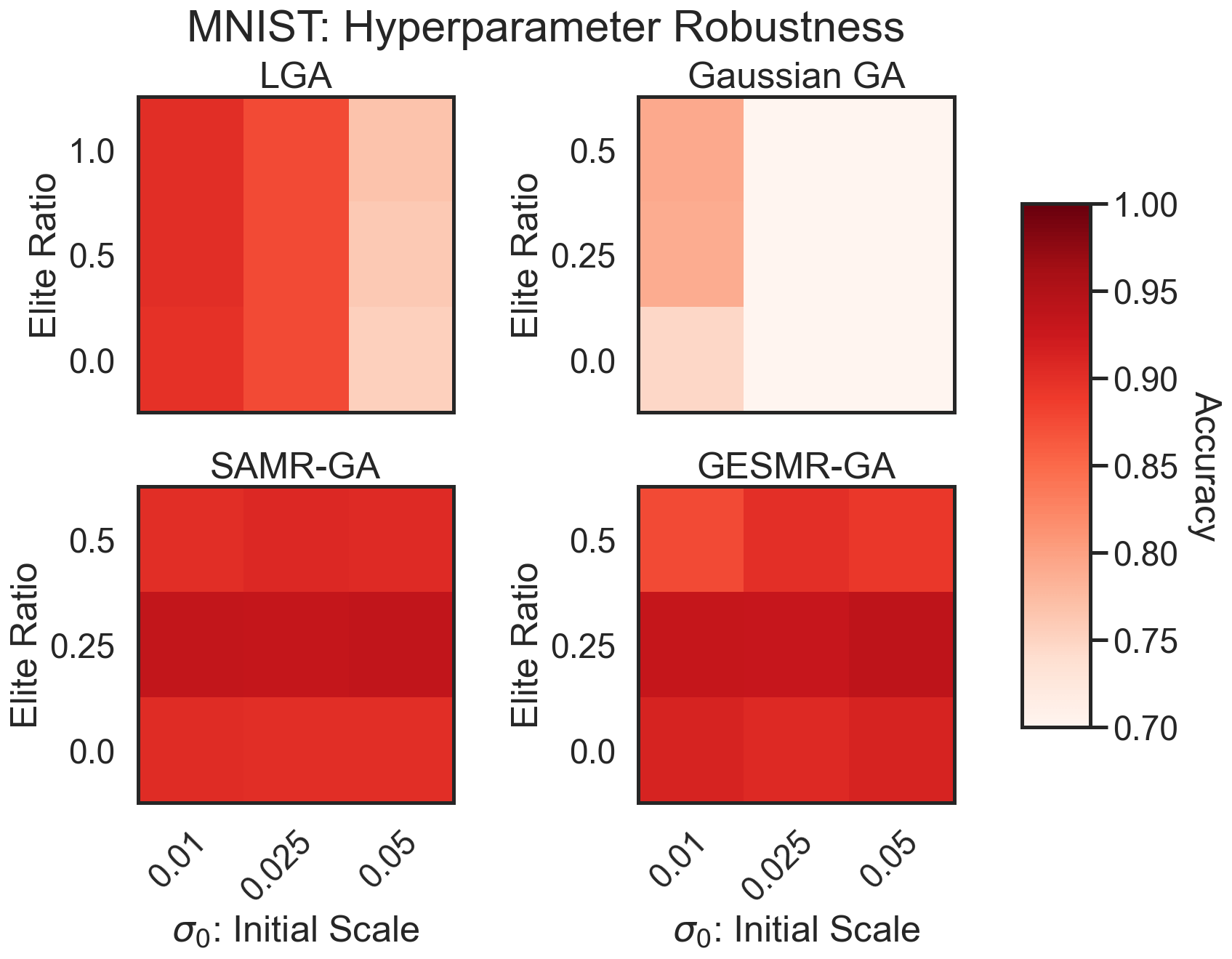}
\end{figure}

\begin{figure}[h]
\caption{Hyperparameter Robustness - F-MNIST Task.}
\centering
\includegraphics[width=0.4\textwidth]{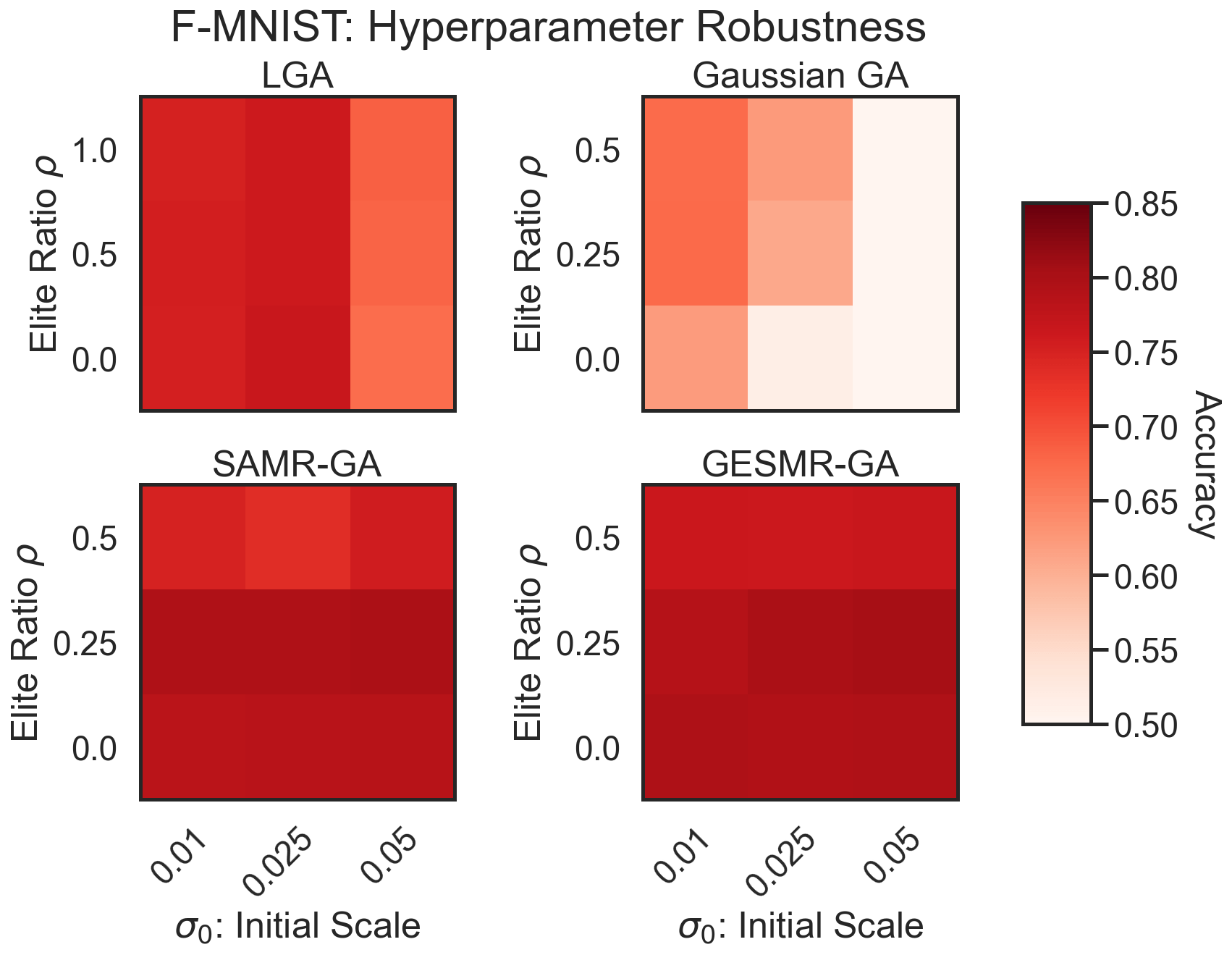}
\end{figure}

\begin{figure}[h]
\caption{Hyperparameter Robustness - K-MNIST Task.}
\centering
\includegraphics[width=0.4\textwidth]{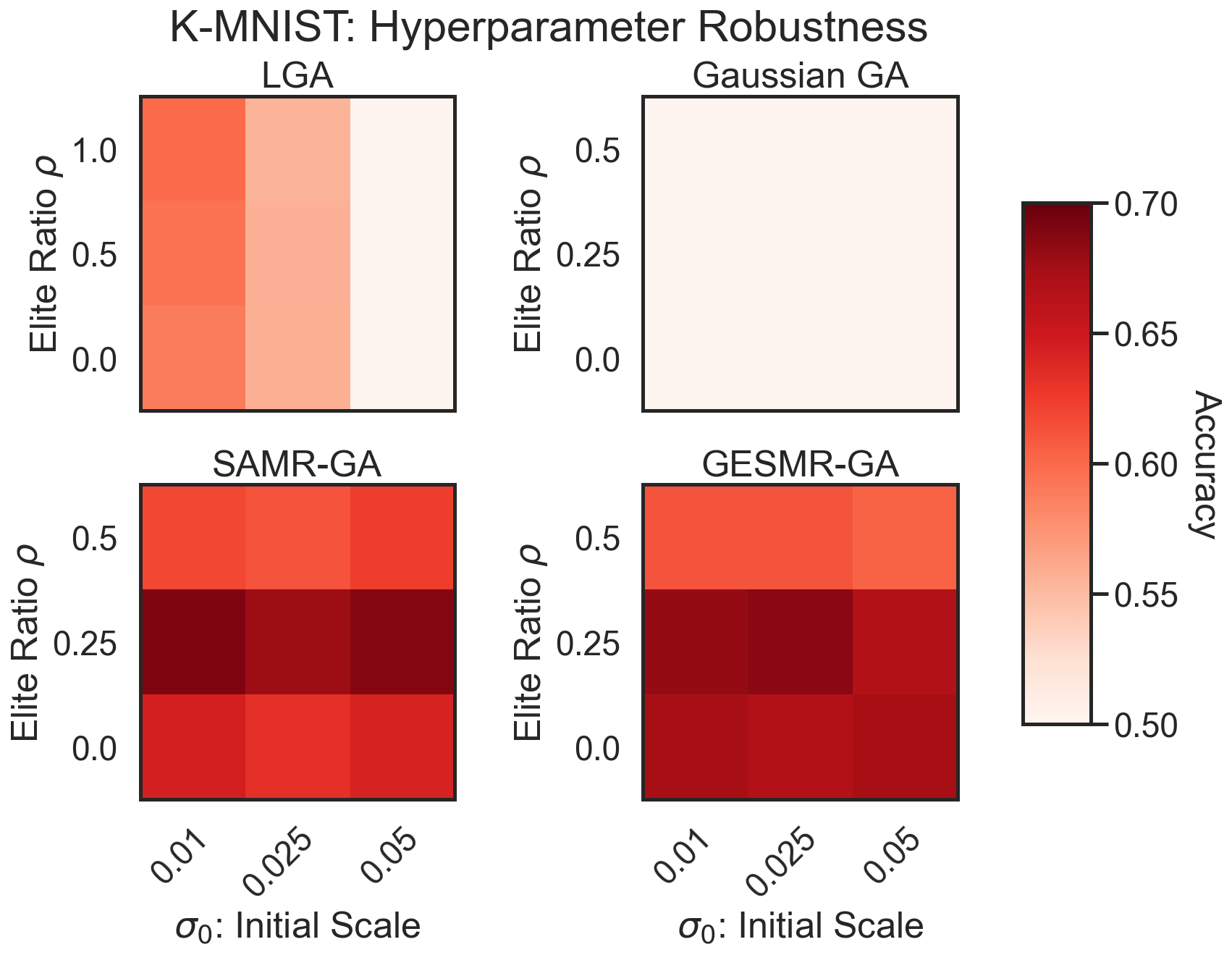}
\end{figure}

\begin{figure}[h]
\caption{Hyperparameter Robustness - Breakout Task.}
\centering
\includegraphics[width=0.4\textwidth]{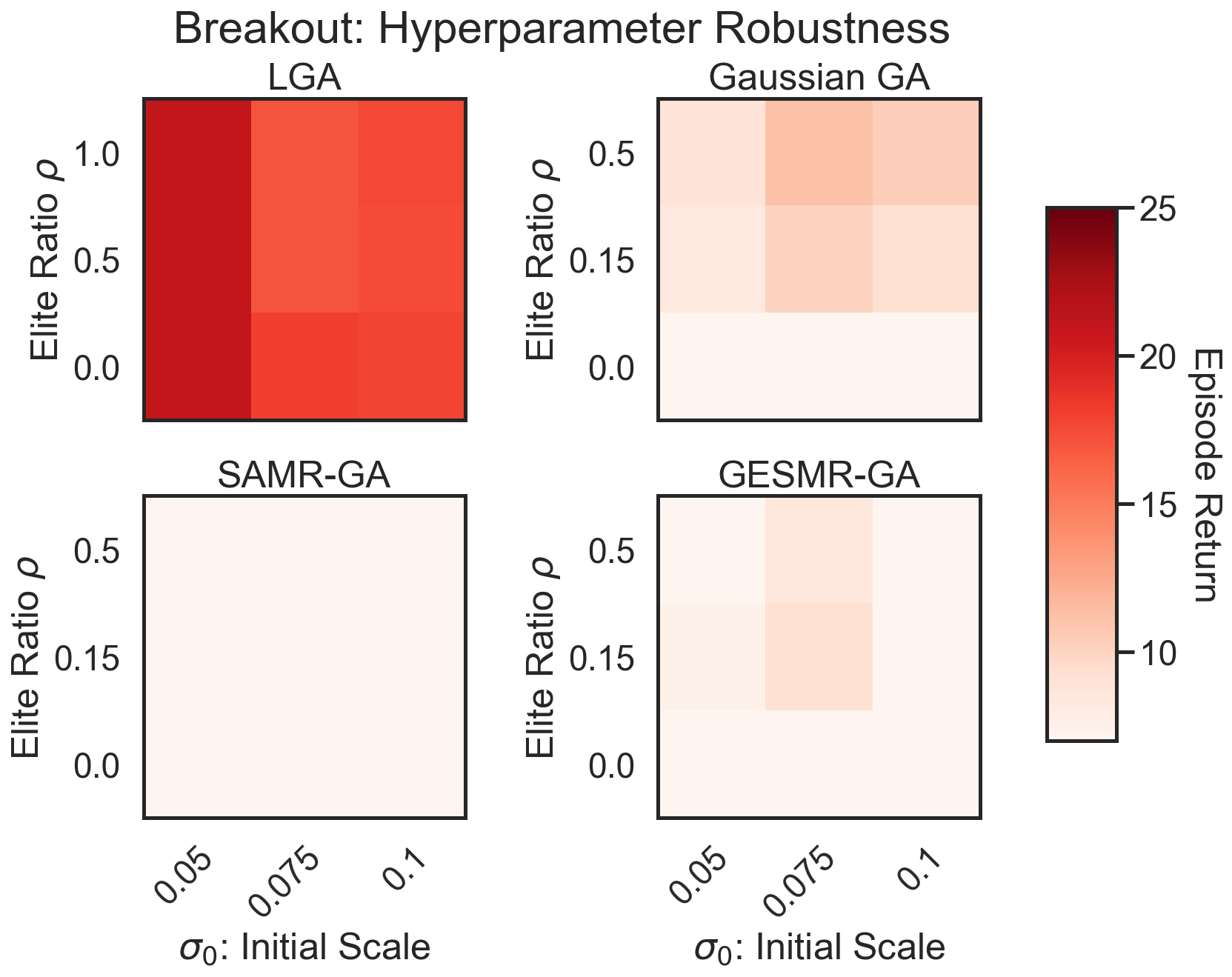}
\end{figure}

\begin{figure}[h]
\caption{Hyperparameter Robustness - Asterix Task.}
\centering
\includegraphics[width=0.4\textwidth]{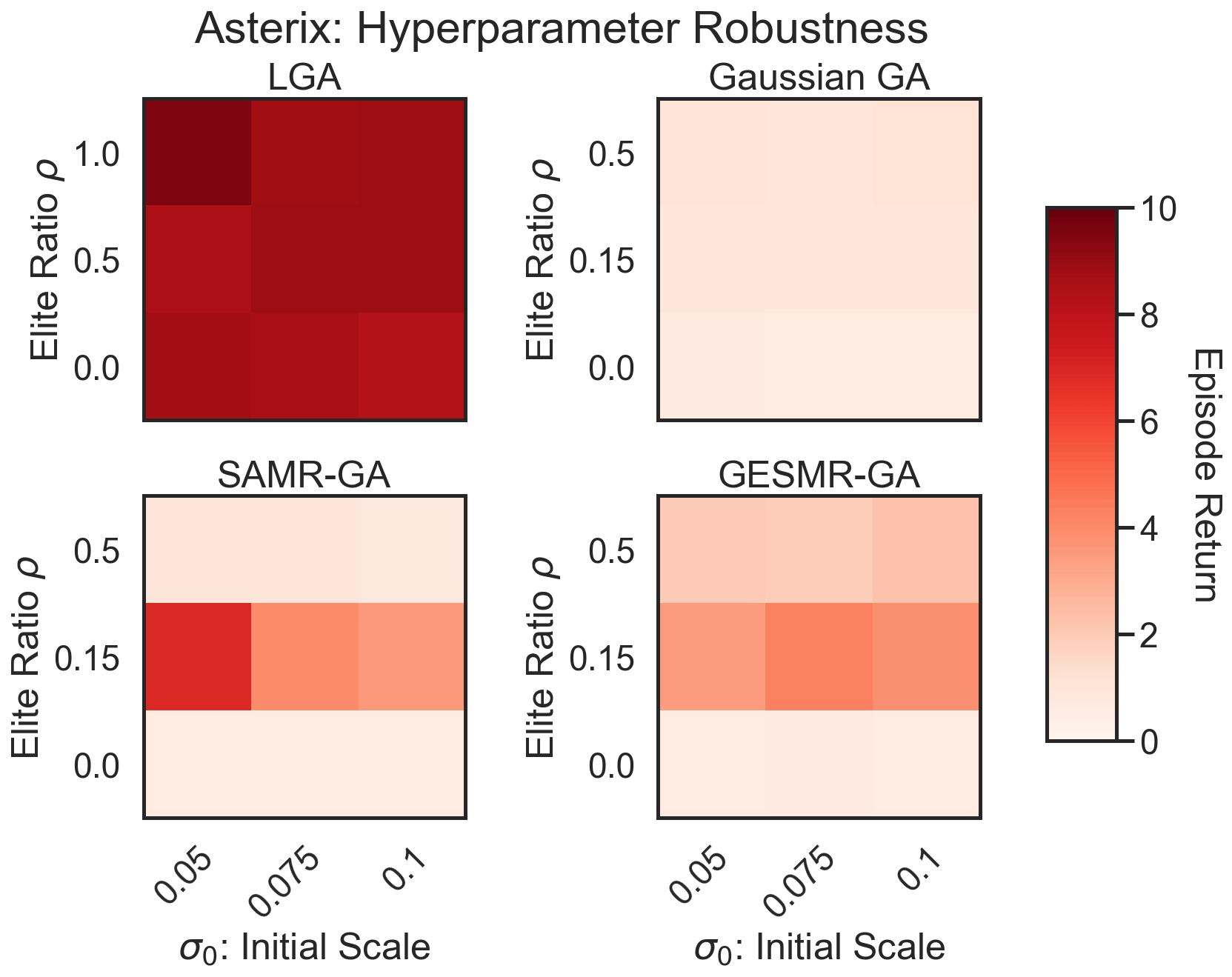}
\end{figure}

\begin{figure}[h]
\caption{Hyperparameter Robustness - SpaceInvaders Task.}
\centering
\includegraphics[width=0.4\textwidth]{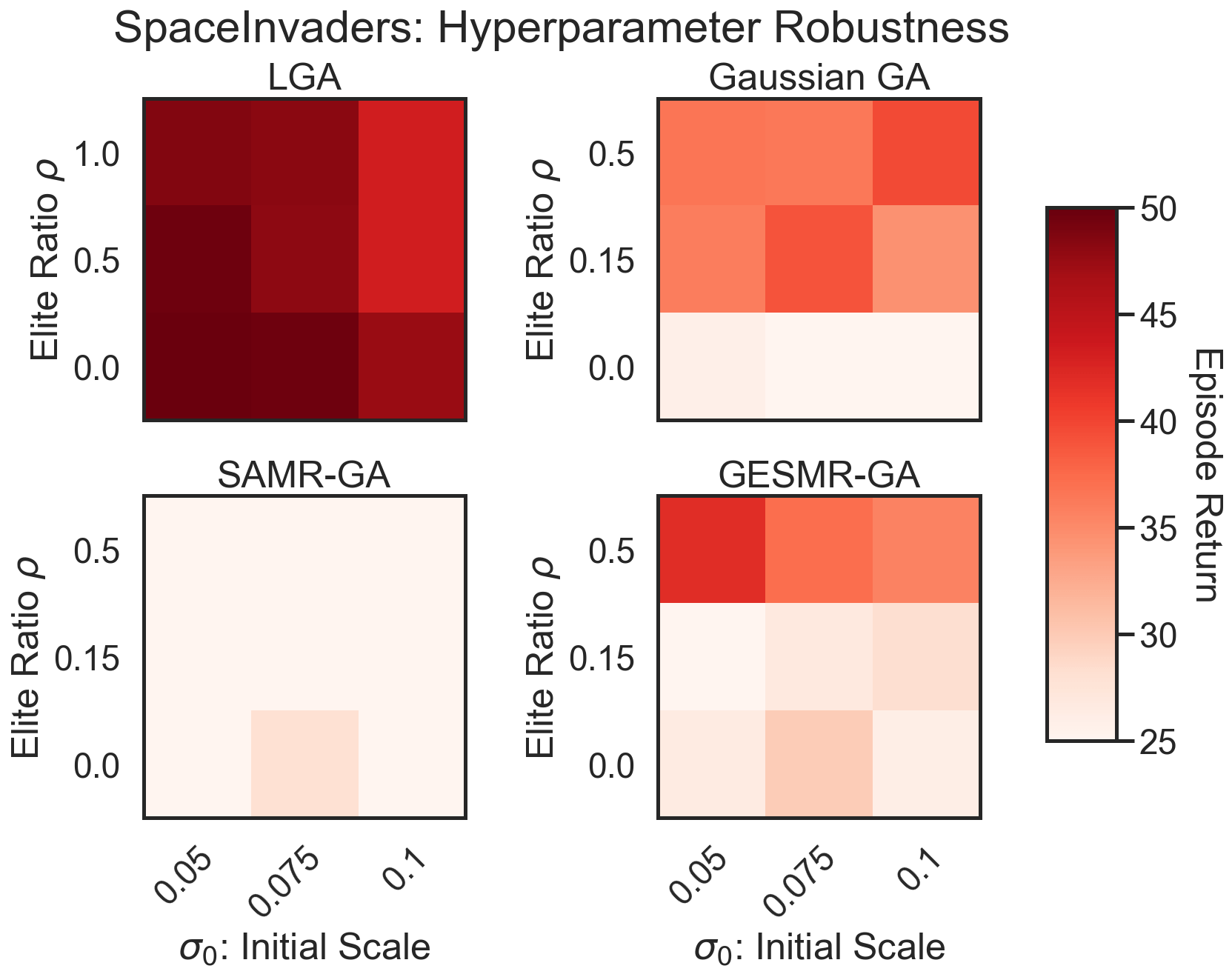}
\end{figure}

\begin{figure}[h]
\caption{Hyperparameter Robustness - Hopper Task.}
\centering
\includegraphics[width=0.4\textwidth]{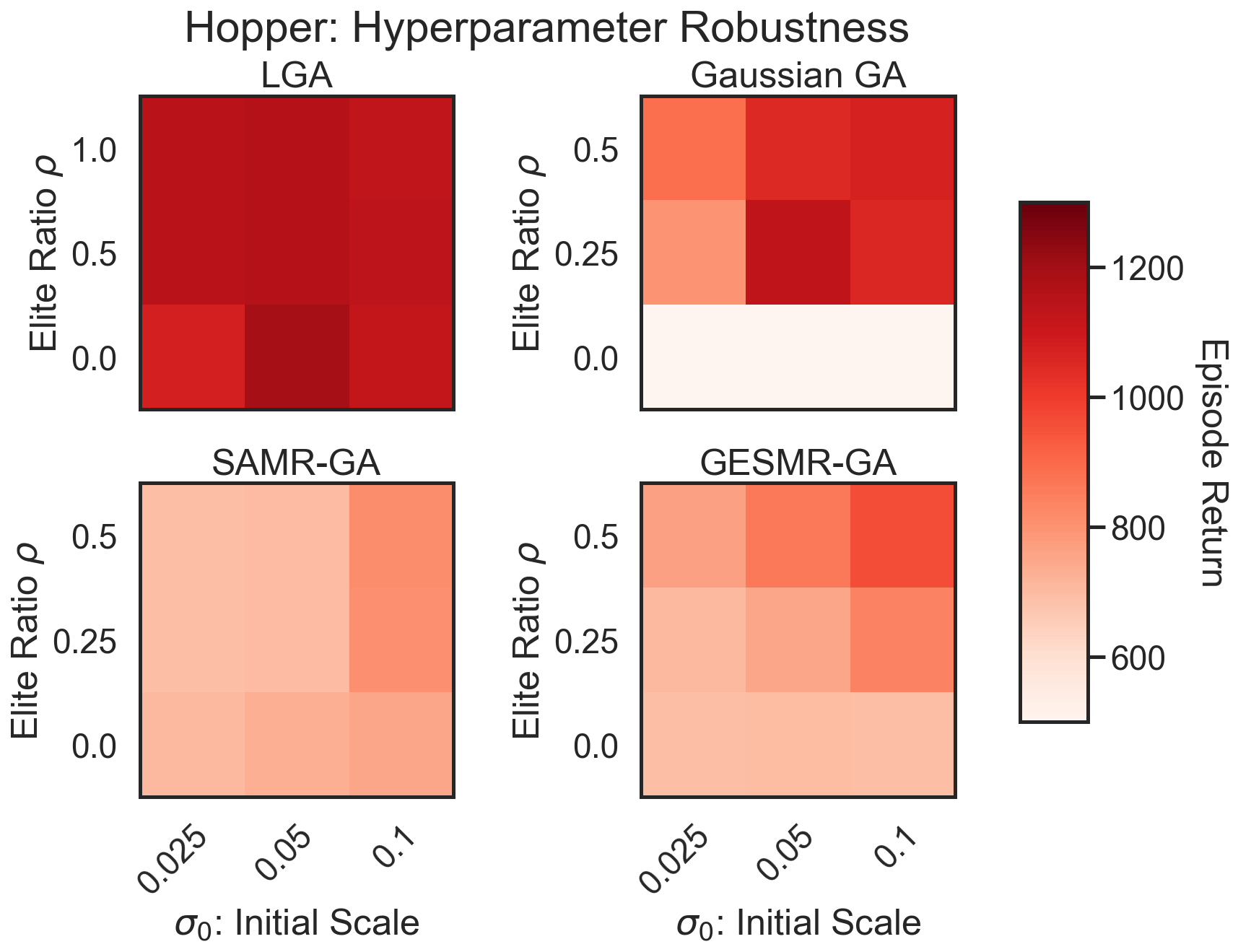}
\end{figure}

\begin{figure}[h]
\caption{Hyperparameter Robustness - HalfCheetah Task.}
\centering
\includegraphics[width=0.4\textwidth]{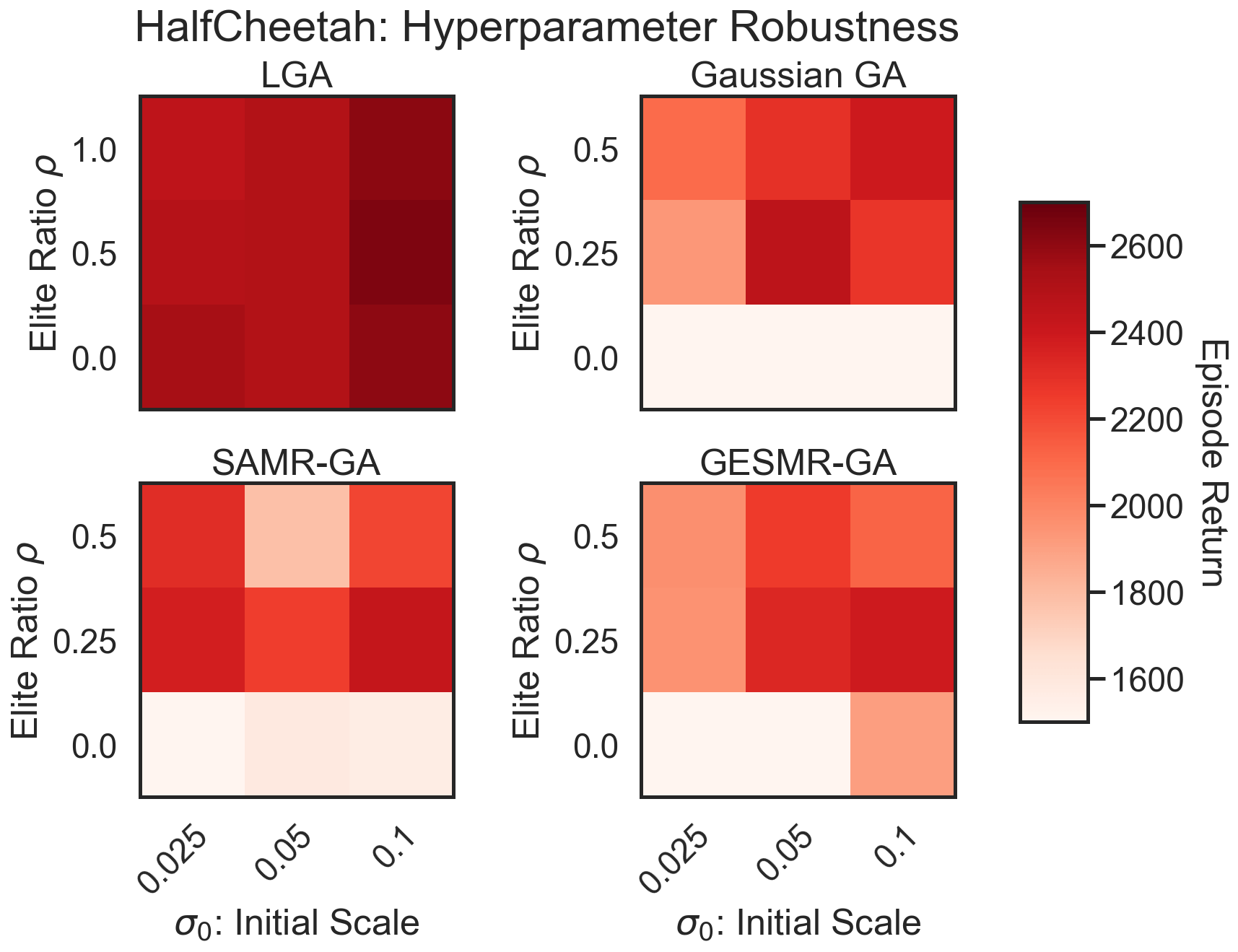}
\end{figure}

\begin{figure}[h]
\caption{Hyperparameter Robustness - Fetch Task.}
\centering
\includegraphics[width=0.4\textwidth]{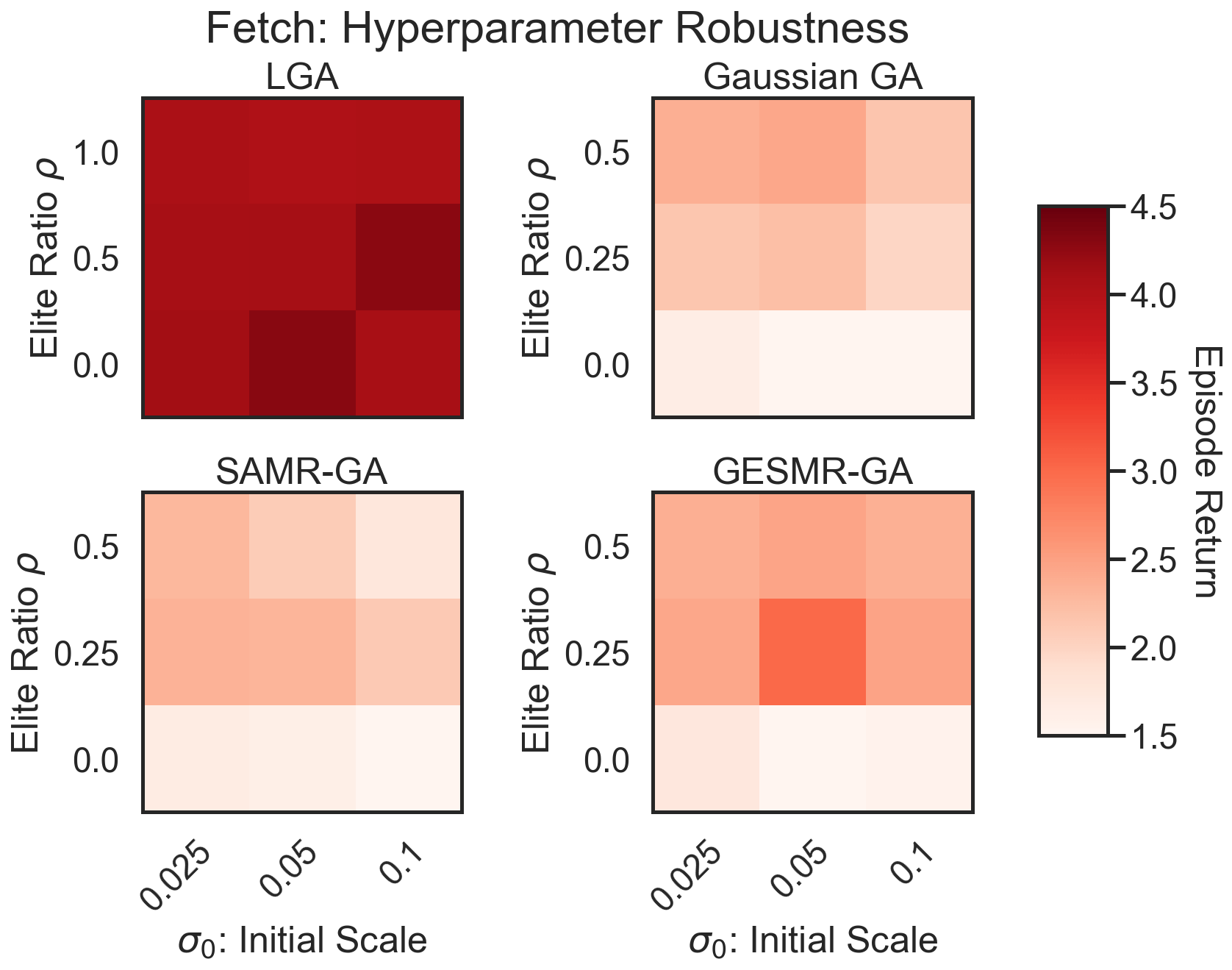}
\end{figure}

\begin{figure}[h]
\caption{Hyperparameter Robustness - Walker2d Task.}
\centering
\includegraphics[width=0.4\textwidth]{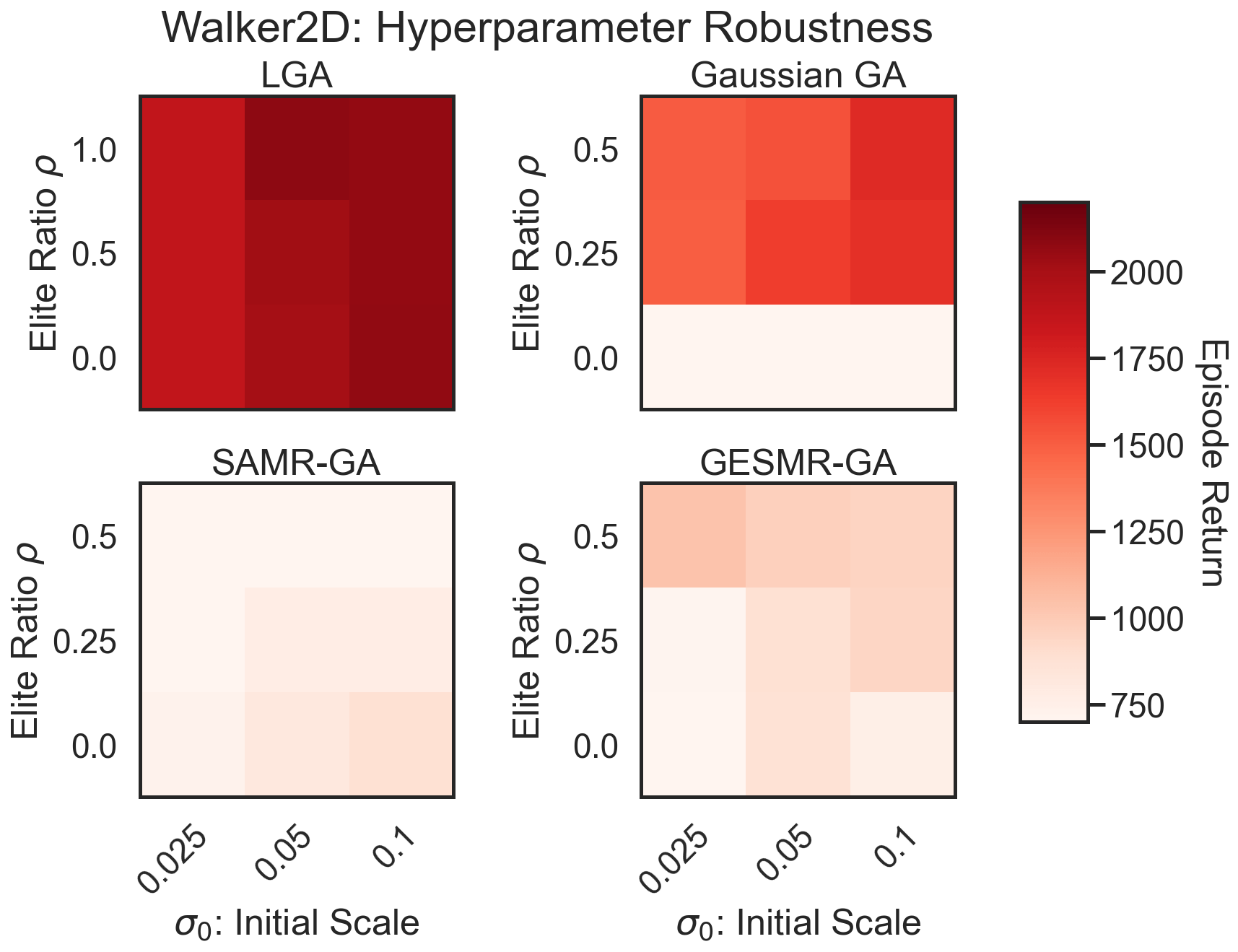}
\end{figure}

\begin{figure}[h]
\caption{Hyperparameter Robustness - ur5e Task.}
\centering
\includegraphics[width=0.4\textwidth]{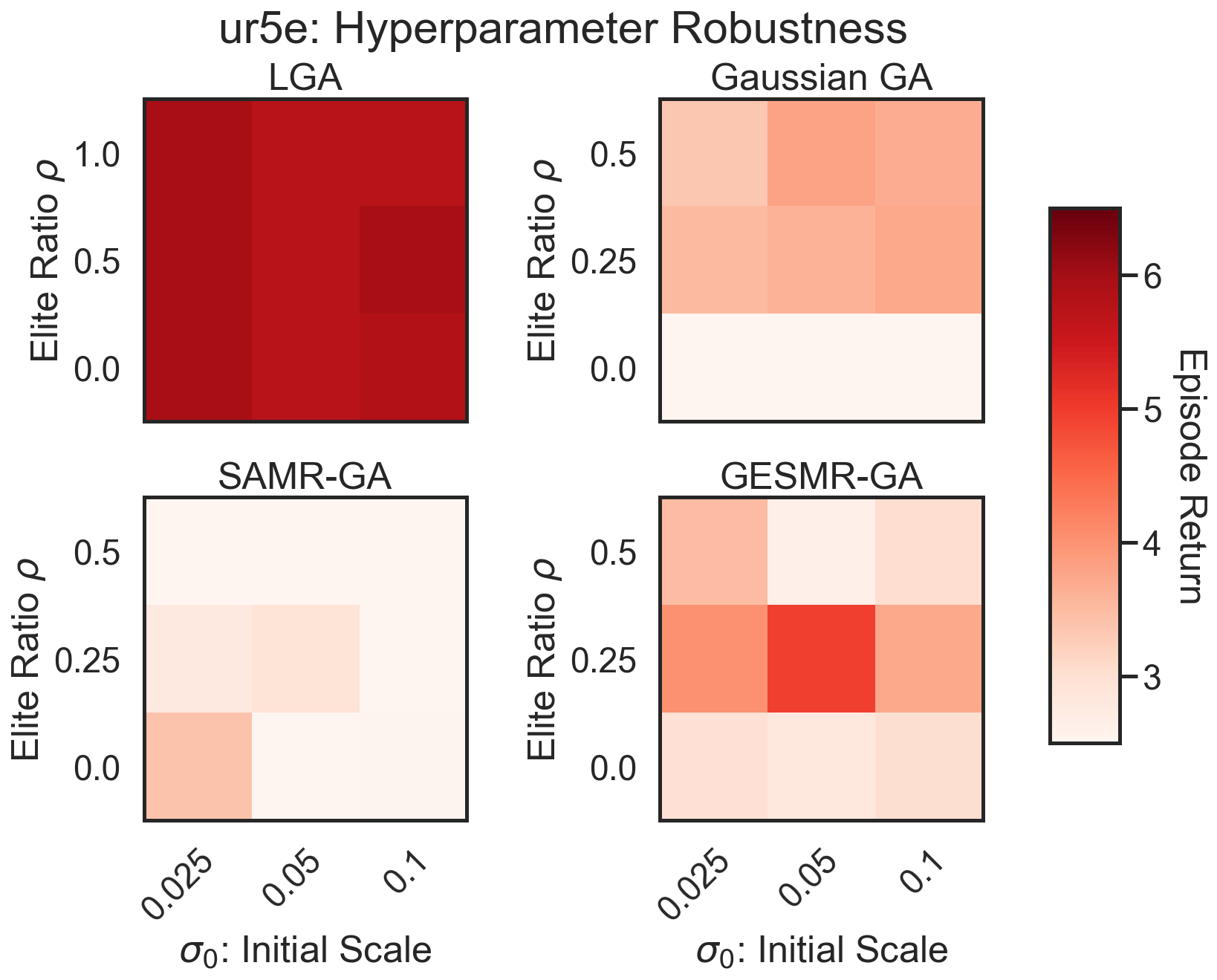}
\end{figure}

\begin{figure}[h]
\caption{Hyperparameter Robustness - Pendulum-v1 Task.}
\centering
\includegraphics[width=0.4\textwidth]{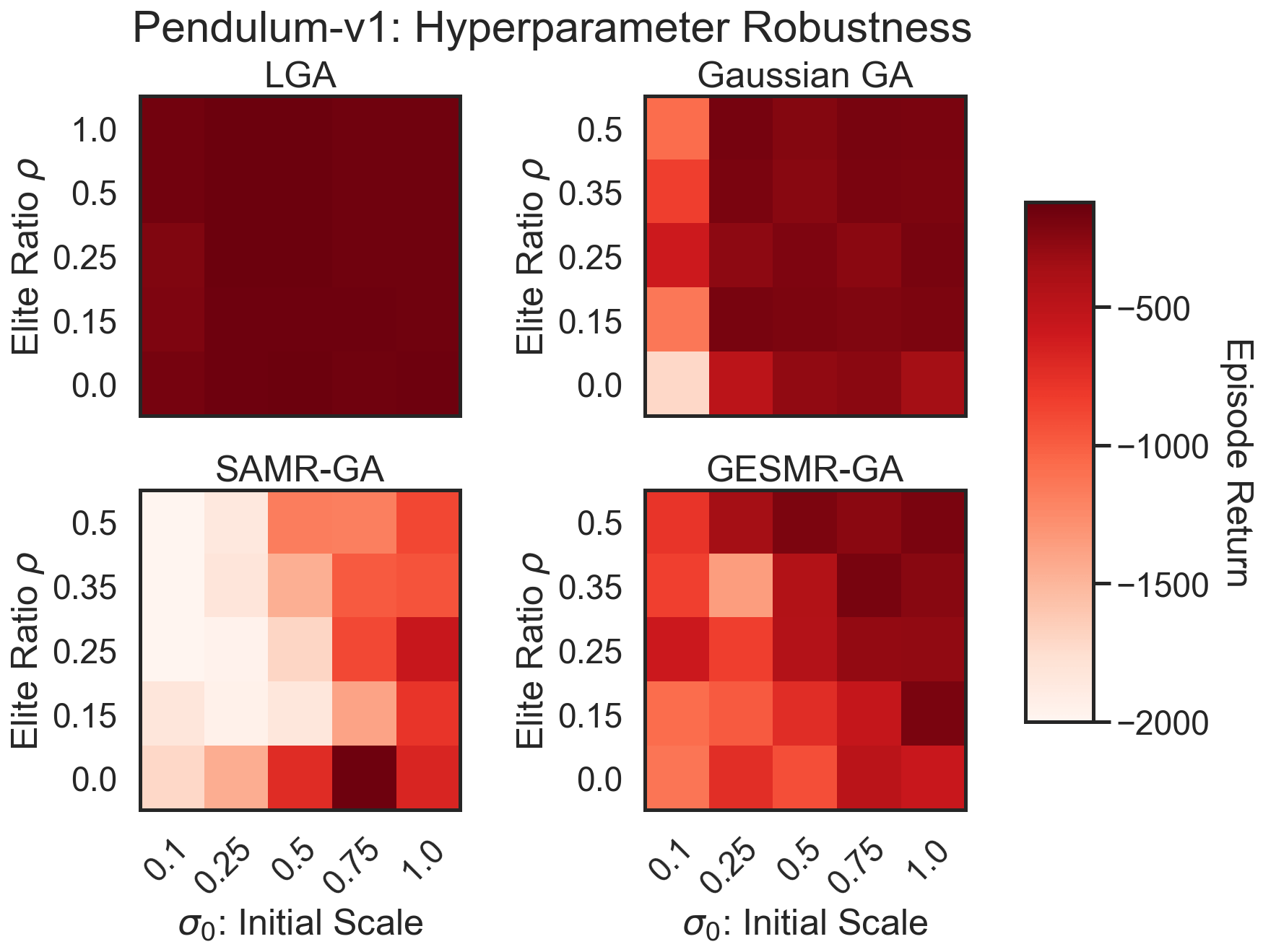}
\end{figure}

\newpage
\section{Reverse-Engineering the Learned GA}
\label{appendix:dga}
\begin{figure}[h]
\caption{\underline{Top}: Linear relationship between fitness features and MRA. \underline{Bottom}: Linear relationship between fitness features and selection logits.}
\centering
\includegraphics[width=0.45\textwidth]{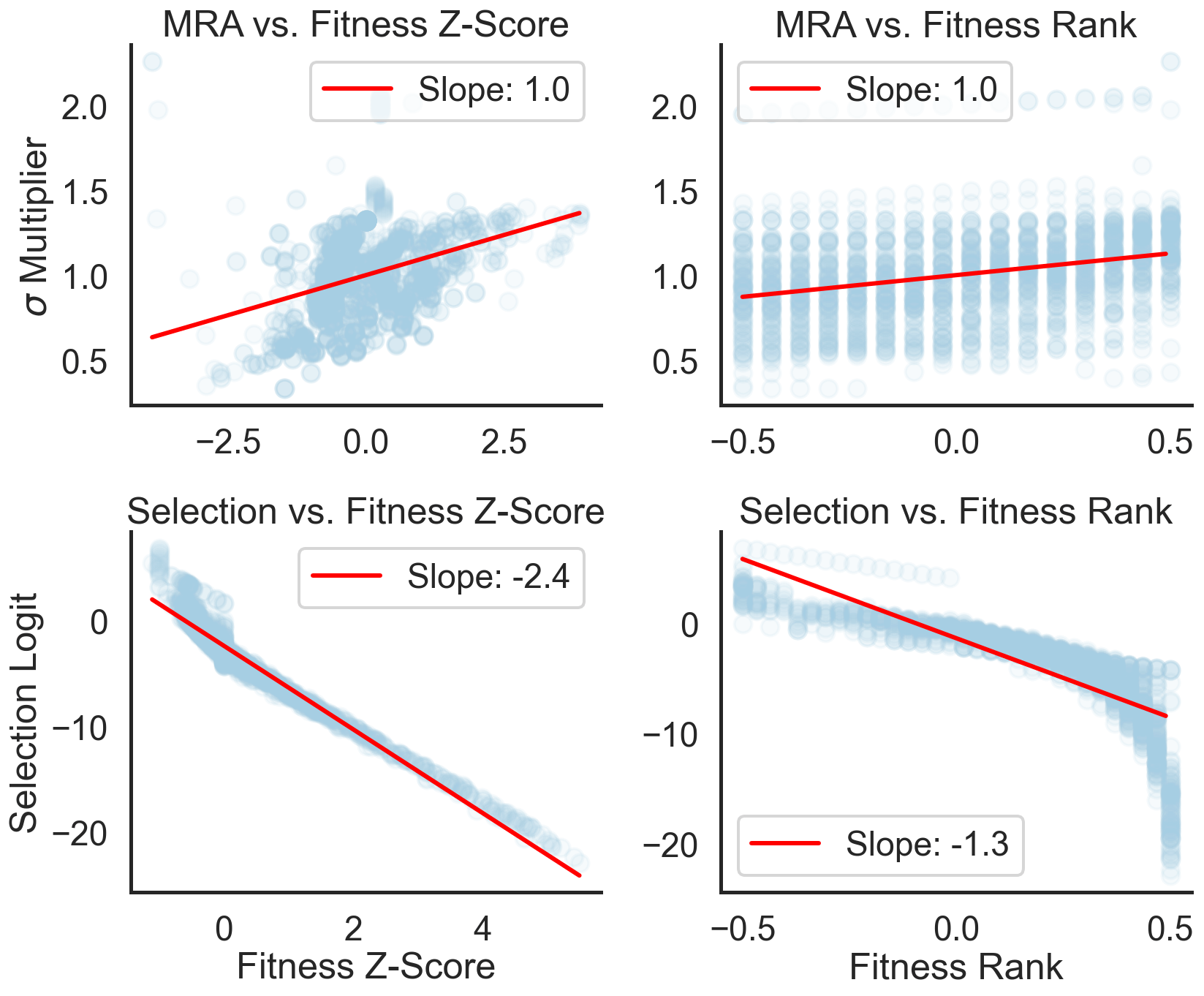}
\label{fig:dga}
\end{figure}

We further investigated whether there exist simple relationships between the attention input features and the learned operator's output. More specifically, we explored if the mutation rate adaption multiplier $\Delta_\sigma$ and selection logits $m_{ij}$ can be explained by the normalized fitness features (z-score and centered ranks) of the children. In Figure \ref{fig:dga} we plot all features, logits and mutation multipliers across an LGA evaluation run on a 2-dim Sphere task. We can observe a clear positive correlation between the performance of the children and their selection probability. Furthermore, the mutation rate is decreased for well-performing solutions. This may open up the possibility to reverse-engineer a new discovered GA without the need for arguably opaque neural network modules.

\section{Software, Compute Requirements}
\label{appendix:software}

This project has been enabled by the usage of freely available Open Source software. This includes the following:

\begin{itemize}
    \item NumPy: \citet{harris2020array}
    \item Matplotlib: \citet{hunter2007matplotlib}
    \item Seaborn: \citet{waskom2021seaborn}
    \item JAX: \citet{jax2018github}
    \item Evosax: \citet{evosax2022github}
    \item Gymnax: \citet{gymnax2022github}
    \item MinAtar: \citet{young2019minatar}
    \item Evojax: \citet{evojax2022}
    \item Brax: \citet{freeman2021brax}
    \item MLE-Infrastructure: \citet{mle_infrastructure2021github}
\end{itemize}

A network checkpoint and accompanying architecture code is publicly available in \href{https://github.com/RobertTLange/evosax}{\texttt{evosax}}.
All experiments (both meta-training and evaluation) were implemented using the JAX library for parallelization of fitness rollout evaluations. Each MetaBBO meta-training was run on 4 RTX2080Ti Nvidia GPUs and take roughly 2.5 hours. 
The LGA downstream BBOB, HPO-B and gym task evaluations were run on a CPU cluster using 2 CPU cores. They last between 2 and 5 minutes. Finally, the neuroevolution tasks were run on individual NVIDIA V100S and A100 GPUs.
The Brax evaluations require between 30 minutes and 1.5 hours depending on the control task. The computer vision evaluation experiments take ca. 10 minutes and the MinAtar experiments last for ca. 1 hour on a V100S GPU.

\end{document}